%% file: Main.tex
\documentclass[sigconf]{acmart}

\AtBeginDocument{%
  \providecommand\BibTeX{{%
    \normalfont B\kern-0.5em{\scshape i\kern-0.25em b}\kern-0.8em\TeX}}}

\copyrightyear{2024}
\acmYear{2024}

\setcopyright{acmlicensed}
\acmConference[KDD '24]{Proceedings of the 30th ACM SIGKDD Conference on Knowledge Discovery and Data Mining}{August 25--29, 2024}{Barcelona, Spain}
\acmBooktitle{Proceedings of the 30th ACM SIGKDD Conference on Knowledge
Discovery and Data Mining (KDD '24), August 25--29, 2024, Barcelona, Spain}
\acmDOI{10.1145/3637528.3671537}
\acmISBN{979-8-4007-0490-1/24/08}

\usepackage{color}
\usepackage{balance}
\usepackage[marginal]{footmisc}

\usepackage{multirow}
\usepackage{ulem}
\useunder{\uline}{\ul}{}
\usepackage{xcolor}

\begin{document}

\title{Xinyu: An Efficient LLM-based System for Commentary Generation}



\author{Yiquan Wu\footnotemark[1]}
\email{wuyiquan@zju.edu.cn}
\affiliation{%
  \institution{Zhejiang University}
  \city{Hangzhou}
  \country{China}
}

\author{Bo Tang\footnotemark[1]}
\email{tangb@iaar.ac.cn}
\affiliation{%
  \institution{University of Science and Technology of China, Hefei, China}
  \institution{Institute for Advanced Algorithms Research, Shanghai, China}
  \country{}
}

\author{Chenyang Xi\footnotemark[2]}
\email{xicy@iaar.ac.cn}
\affiliation{%
  \institution{Institute for Advanced Algorithms Research}
  \city{Shanghai}
  \country{China}
}

\author{Yu Yu\footnotemark[2]}
\email{yuy@iaar.ac.cn}
\affiliation{%
  \institution{Institute for Advanced Algorithms Research}
  \city{Shanghai}
  \country{China}
}

\author{Pengyu Wang}
\email{2371407@stu.neu.edu.cn}
\affiliation{%
  \institution{Northeastern University}
  \city{Shenyang}
  \country{China}
}

\author{Yifei Liu}
\email{liuyifei@zju.edu.cn}
\affiliation{%
  \institution{Zhejiang University}
  \city{Hangzhou}
  \country{China}
}

\author{Kun Kuang\footnotemark[3]}
\email{kunkuang@zju.edu.cn}
\affiliation{%
  \institution{Zhejiang University}
  \city{Hangzhou}
  \country{China}
}

\author{Haiying Deng}
\email{denghaiying@xinhuaskl.com}
\affiliation{%
  \institution{State Key Laboratory of Media Convergence Production Technology and Systems}
  \city{Beijing}
  \country{China}
}

\author{Zhiyu Li}
\email{lizy@iaar.ac.cn}
\affiliation{%
  \institution{Institute for Advanced Algorithms Research}
  \city{Shanghai}
  \country{China}
}

\author{Feiyu Xiong}
\email{xiongfy@iaar.ac.cn}
\affiliation{%
  \institution{Institute for Advanced Algorithms Research}
  \city{Shanghai}
  \country{China}
}

\author{Jie Hu}
\email{hujie1@chinatelecom.cn}
\affiliation{%
  \institution{Research Institute of China Telecom}
  \city{Beijing}
  \country{China}
}

\author{Peng Cheng}
\email{chengpeng@xinhua.org}
\affiliation{%
  \institution{State Key Laboratory of Media Convergence Production Technology and Systems}
  \city{Beijing}
  \country{China}
}

\author{Zhonghao Wang}
\email{wangzhonghao@xinhua.org}
\affiliation{%
  \institution{State Key Laboratory of Media Convergence Production Technology and Systems}
  \city{Beijing}
  \country{China}
}

\author{Yi Wang}
\email{wangyi08@xinhua.org}
\affiliation{%
  \institution{State Key Laboratory of Media Convergence Production Technology and Systems}
  \city{Beijing}
  \country{China}
}

\author{Yi Luo}
\email{luoyi@xinhua.org}
\affiliation{%
  \institution{State Key Laboratory of Media Convergence Production Technology and Systems}
  \city{Beijing}
  \country{China}
}

\author{Mingchuan Yang}
\email{yangmch@chinatelecom.cn}
\affiliation{%
  \institution{Research Institute of China Telecom}
  \city{Beijing}
  \country{China}
}



\renewcommand{\shortauthors}{Yiquan Wu et al.}


\begin{abstract}
Commentary provides readers with a deep understanding of events by presenting diverse arguments and evidence. However, creating commentary is a time-consuming task, even for skilled commentators.
Large language models (LLMs) have simplified the process of natural language generation, but their direct application in commentary creation still faces challenges due to unique task requirements. These requirements can be categorized into two levels: 1) fundamental requirements, which include creating well-structured and logically consistent narratives, and 2) advanced requirements, which involve generating quality arguments and providing convincing evidence.
In this paper, we introduce Xinyu, an efficient LLM-based system designed to assist commentators in generating Chinese commentaries. To meet the fundamental requirements, we deconstruct the generation process into sequential steps, proposing targeted strategies and supervised fine-tuning (SFT) for each step. To address the advanced requirements, we present an argument ranking model for arguments and establish a comprehensive evidence database that includes up-to-date events and classic books, thereby strengthening the substantiation of the evidence with retrieval augmented generation (RAG) technology.
To evaluate the generated commentaries more fairly, corresponding to the two-level requirements, we introduce a comprehensive evaluation metric that considers five distinct perspectives in commentary generation.
Our experiments confirm the effectiveness of our proposed system. We also observe a significant increase in the efficiency of commentators in real-world scenarios, with the average time spent on creating a commentary dropping from 4 hours to 20 minutes. Importantly, such an increase in efficiency does not compromise the quality of the commentaries.
\end{abstract}






\maketitle

\footnotetext[1]{* These authors contributed equally to this work.}
\footnotetext[2]{† These authors contributed equally to this work.}
\footnotetext[3]{‡ Corresponding Author.}

\input{1_Introduction}
\input{2_Related_Work}
\input{3_Preliminaries}

\input{4_Method}

\input{5_Experiment}
\input{6_Conclusion}
\input{Acknowledgment}

\bibliographystyle{ACM-Reference-Format}
\balance
\bibliography{sample-base}

\input{7_Appendix}

\end{document}

%% file: 1_Introduction.tex
\section{Introduction}


\begin{figure*}[t]
\centering
\includegraphics[width=0.9\textwidth]{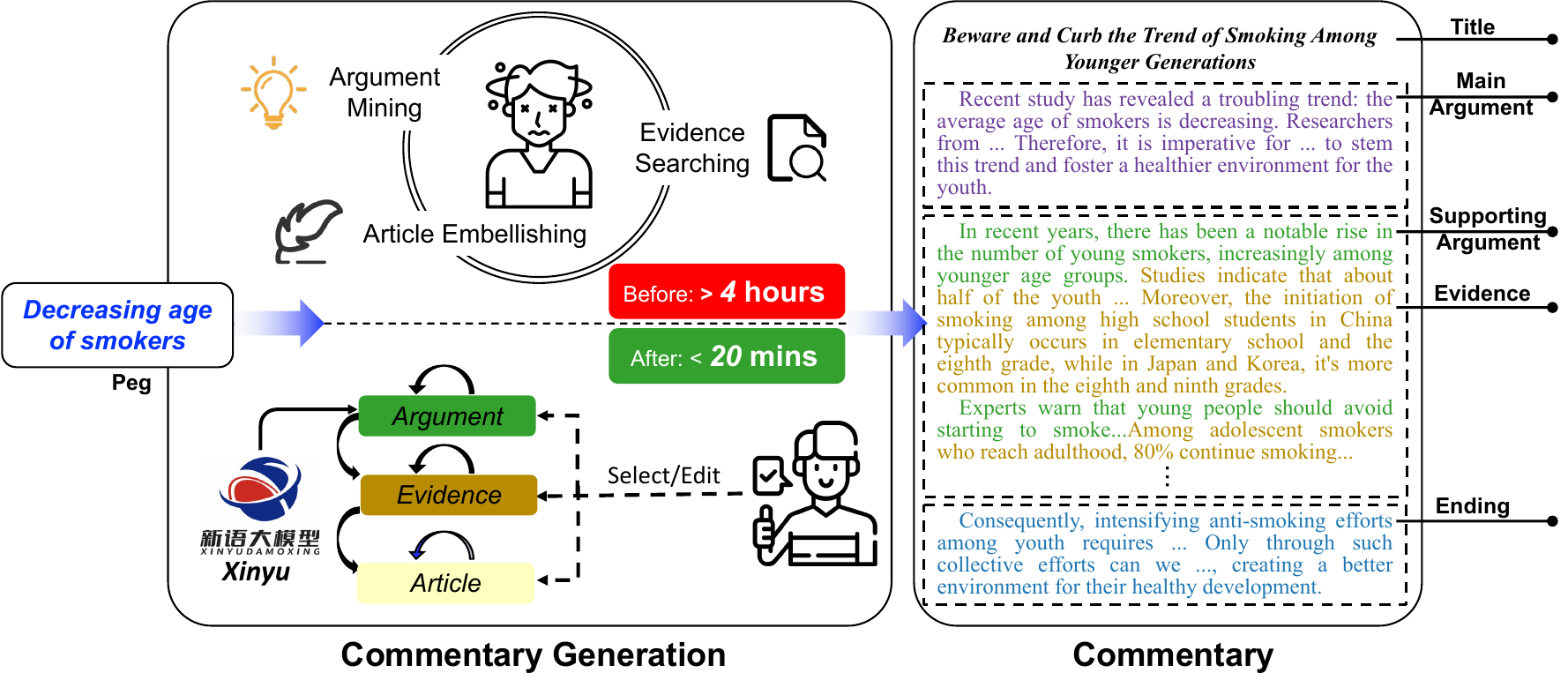}
\caption{
The illustration of the commentary generation task. To generate a commentary, it usually requires argument mining, evidence searching, and article embellishing. With the Xinyu, the intermediate steps can be sped up. The right part demonstrates the structure of a commentary, which consists of a title, a main argument, several supporting arguments and evidence, and an ending. This example is translated from Chinese.
}
\label{fig:example}
\end{figure*}

With the advancement of natural language processing (NLP), particularly large language models (LLMs), numerous text-generation systems have been proposed to enhance the effectiveness and efficiency of individuals across various fields, such as education, medicine and law \citep{adeshola2023opportunities, wu2023precedent, cheng2023now}. Commentary is a type of article that contains diverse arguments and compelling evidence, which aims to provide readers with a deep understanding of certain events. As Figure. \ref{fig:example} shows, a commentator usually spends several hours writing a commentary, which includes mining arguments, searching for evidence, and embellishing the article. Given the continuous nature of news, their workload is substantial. Therefore, exploring the application of LLMs in commentary generation is worthwhile.

Although LLMs have benefited many generative tasks, they face challenges when directly applied to commentary generation due to unique task requirements. Broadly, the requirements for a commentary can be divided into two levels:

\textbf{1) Fundamental requirements:} 

$\bullet$ The structure should be regular and complete. As Figure. \ref{fig:example} shows, the commentary should follow a total division structure.

$\bullet$ The content should be self-consistent. For example, the arguments in the commentary should not be contradictory, and the evidence must support the arguments.

\textbf{2) Advanced requirements:} 

$\bullet$ Arguments should be specific and original. The argument is key to the commentary, representing the author's stance.

$\bullet$ Evidence should be convincing, which means the LLMs can't generate fake evidence, and the evidence is preferably new.

%

In this paper, we propose Xinyu, an efficient LLM-based system to assist commentators in Chinese commentary generation. Specifically, for the fundamental requirements, we decompose the generation into several sequential steps, ensuring the generated text is well-structured. We also design targeted supervised fine-tuning (SFT) and strategies for each step to maintain content consistency. For the advanced requirements, we propose an argument ranking model for ranking candidate arguments to ensure quality. Moreover, we construct a comprehensive evidence database, which maintains up-to-date events and books, and then use the technology of retrieval augmented generation (RAG) to generate convincing evidence.

Given the dynamic nature of commentary, traditional metrics for text generation tasks, such as ROUGE or BLEU, fall short in evaluating the overall quality of the commentary. Thus, we propose a comprehensive of evaluation metric for commentary generation that considers five distinct perspectives. In our pilot study, GPT-4 demonstrated performance on par with human annotators, so we employ GPT-4 as the evaluator. The experimental results underscore the quality of the content generated by our system, Xinyu.
In practical terms, we also examined how Xinyu could enhance the work efficiency of human commentators and the result shows that with Xinyu, the speed of commentary generation increased dramatically, reducing the average creation time from 4 hours to a mere 20 minutes. Importantly, this increase in efficiency does not sacrifice the quality of the commentaries.

To sum up, our main contributions are as follows:
\begin{itemize}
    \item We leverage LLMs for the task of commentary generation and propose a system named Xinyu that can assist commentators in generating Chinese commentary 10 times faster with even quality.

    \item We decompose the commentary and generate it in steps, applying targeted supervised fine-tuning (SFT) for each. This approach ensures the commentary meets its fundamental requirements: it is well-structured and self-consistent.

    \item We propose an argument ranking module to improve the quality of the arguments and construct a comprehensive knowledge database (e.g., up-to-date events and books) for the generation of evidence with the help of retrieval augmented generation (RAG). This approach ensures the commentary meets the advanced requirements: it is specific and convincing.

    \item We design a comprehensive evaluation method for the commentary generation task with 5 distinct perspectives. The experimental results demonstrate the effectiveness of our proposed Xinyu.

\end{itemize}

%% file: 2_Related_Work.tex
\section{Related Work}



\subsection{Large Language Models}

The domain of Natural Language Processing (NLP) has witnessed substantial progress \citep{wu2020biased, shen2022mask, zhou2022similar, wu2023focus, liu2023ml}, especially through the advent of Large Language Models (LLMs) \citep{instructgpt,gpt4, llama, baichuan,qwen}. These models show exceptional text generation proficiency, yielding high fluency and readability outputs \citep{wu2023precedent, zhang2024plad}. Their ability to adapt to downstream tasks with minimal in-context examples is particularly noteworthy. To further augment the efficacy of LLMs in downstream tasks, two main methods have been identified: supervised fine-Tuning (SFT) and retrieval augmented generation (RAG).

\textbf{Supervised Fine-Tuning} (SFT) entails the adaptation of an LLM to a specific downstream task. This process refines the model's parameters to align with the data distribution and task requirements, ensuring the model's behavior mirrors human behavior within the given domain. The topic of SFT has been extensively explored in numerous research.
\citet{instructgpt} pioneered the introduction of supervised fine-tuning and reinforcement learning to align language models with human intent.
\citet{lima} compiled a dataset of merely 1K examples for SFT, demonstrating that the success of SFT depends on the quality and diversity of data.

\textbf{Retrieval Augmented Generation} (RAG) amalgamates LLMs with content retrieved from external databases. This approach offers a promising solution to the challenges encountered by LLMs, such as hallucination, outdated knowledge, and untraceable reasoning processes. The conventional RAG process encompasses indexing, retrieval, and generation \citep{retrieve-read,rag-survey}.
RAG has been further enhanced by a range of innovative techniques:
fine-tuning retrieval models to obtain precise semantic representations \citep{UAE,VoyageAI, bge_embedding}, 
reformulating queries to align with the semantic space of queries and documents \citep{hyde,wang2023query2doc,shao2023enhancing}, 
fine-tuning LLMs to harmonize the output of the retriever with the LLM's preference \citep{xu2023recomp,izacard2022atlas,shi2023replug}.

In our work, we leverage the advances of both SFT and RAG to enhance the performance of the Xinyu.

\subsection{Domain-specific LLMs}

Large Language Models (LLMs) have advanced the field of natural language processing, providing a task-agnostic foundation for a wide range of applications. However, directly applying LLMs to solve sophisticated problems in specific domains meets many hurdles, caused by the heterogeneity of domain data, the sophistication of domain knowledge, and the diversity of the constraints \cite{domain-survey}.

Numerous researchers have devoted their efforts to domain-specific Language Models (LLMs) tailored for various fields.
These specialized LLMs have been designed to cater to the unique requirements of domains such as 
medicine \citep{chatdoctor,med-llm,med-llm2} for medical diagnosis, 
law \citep{chatlaw, fedjudge,lawllm} for handling legal documents,
counselling \citep{chatcounselor} for mental health support, 
education \citep{edu-gpt,edu-llm} for teaching assistance,
science\citep{science} for crafting scientific journalism, and so on.

The former domain-specific LLMs mainly focus on injecting domain knowledge (e.g., medical or legal knowledge) into LLMs. In this paper, our focus is on commentary generation, to support commentators in their writing process and produce well-structured, logically consistent commentaries that present novel arguments and convincing evidence.

%% file: 3_Preliminaries.tex
\section{Preliminaries}

This section is dedicated to 
defining key concepts that will be consistently referenced throughout this paper.

\noindent \textbf{Peg}, within the scope of commentary generation, denotes the specific aspect of the event that the commentary is responding to or building upon. It acts as an anchor for the commentary. For instance, in Figure. \ref{fig:example}, the peg is `Age of smokers decrease'.

\noindent \textbf{Main Argument}, in the context of commentary generation, signifies the central point that the generated commentary seeks to communicate. It forms the core message around which the commentary is structured.

\noindent \textbf{Supporting Argument}, is an additional point that helps to substantiate the main argument. Typically, a commentary will contain several supporting arguments that collectively contribute to the strength and depth of the main argument.

\noindent \textbf{Evidence} refers to the data, facts, or information employed to support the argument. In the process of commentary generation, evidence can be derived from the content itself or external sources.

Given a peg, the corresponding commentary will include one main argument and several supporting arguments, all of which are supported by evidence.

%% file: 4_Method.tex
\begin{figure*}[t]
\centering
\includegraphics[width=0.9\textwidth]{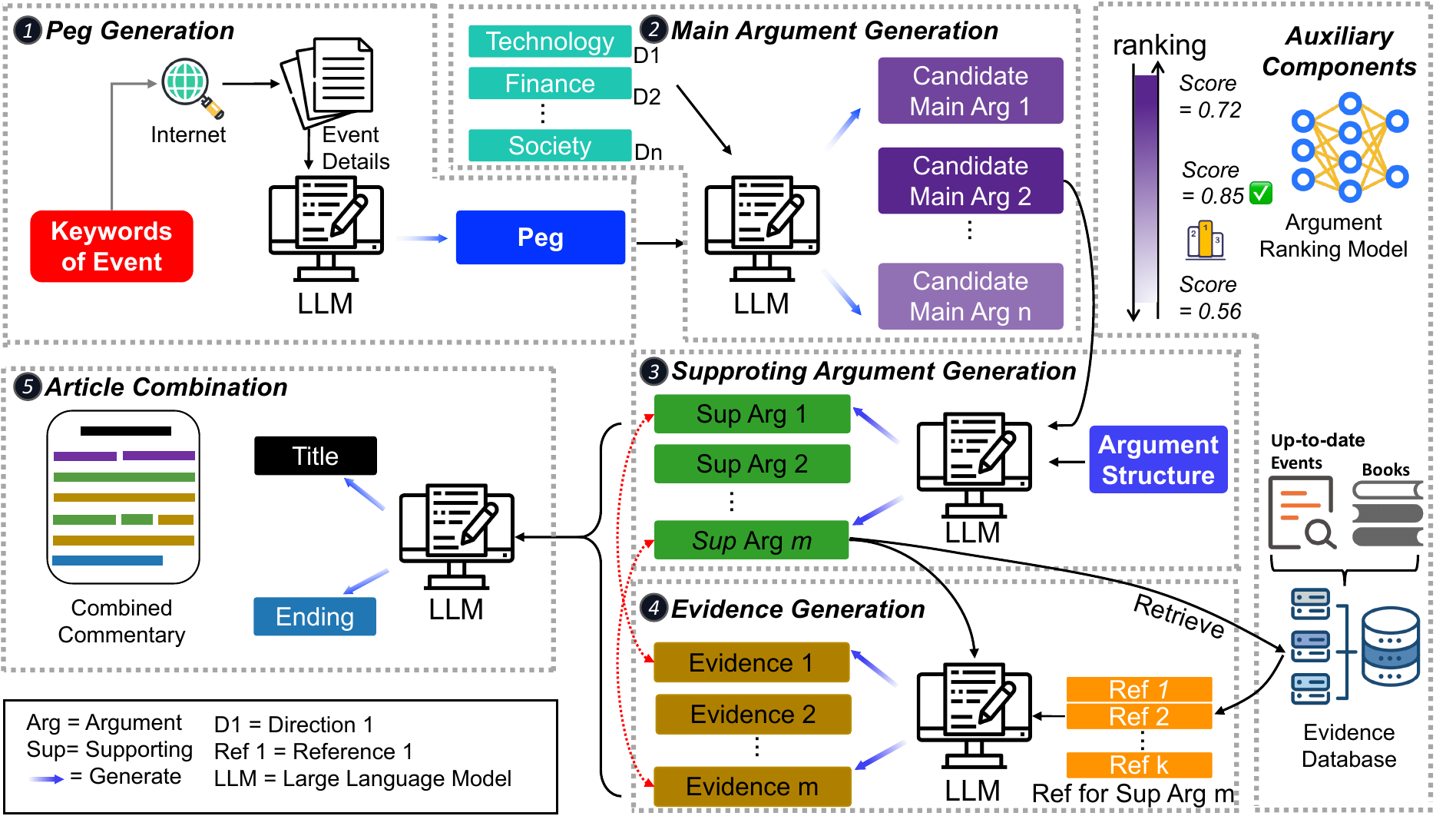}
\caption{The overall framework of Xinyu. The generation process is divided into 5 steps.
}
\label{fig:frame}
\end{figure*}

\section{Technical route of Xinyu}
%
In this section, we delve into Xinyu's comprehensive technical route. Figure. \ref{fig:frame} shows the overall framework of Xinyu. In Section 4.1, we introduce the five main generative components used in detail during the commentary generation process. In Section 4.2, we shift our focus to the two auxiliary components essential for meeting the advanced requirements of the commentary, including the argument ranking model and the construction of an evidence database. Note that without these two auxiliary components, the system can still fulfill the fundamental requirements.

\subsection{Main Components}
Based on the structure, we decompose the commentary generation into five steps: peg generation, main argument generation, supporting argument generation, evidence generation, and finally, article combination. This sequential approach is implemented with the help of SFT and RAG\footnote{Note the user can interact with Xinyu by providing additional input or editing the output at each step and here we mainly describe the automatic process.}.

\subsubsection{Peg Generation}
The peg generation serves as a preliminary step in the commentary generation process, designed to swiftly summarize event details for the user. Utilizing a search engine, this component retrieves event details based on given keywords to generate a peg. Alternatively, users have the option to manually compose the peg, bypassing this automated step.
Specifically, the content from the top three most relevant search results is processed as input, and the LLM condenses this information into a concise peg. The procedure is exemplified as follows:

\textit{[You are a commentary writing expert, and here are the details of an event. Event detail: \{\textbf{event detail}\}. Please refine it into a concise and well-articulated peg:]}

To enhance the model's proficiency in condensing event details during peg generation, we develop SFT data specifically for this step. By inputting event details and using the peg as a label, this method trains the model to more effectively summarize and pinpoint essential information, resulting in pegs that are informationally dense.

\subsubsection{Main Argument Generation}
This step aims to provide the main argument. Due to the variety of pegs, main arguments can be driven in different directions. Here the strategy involves directing the LLM to generate across ten distinct directions: technology, finance, society, politics, literature and arts, lifestyle, environment, sports, education, and science. Each direction emphasizes its specific thematic elements, such as highlighting technological advancements or economic trends.
To operationalize this strategy, we combine the peg, event details, and a chosen direction as input for the LLM, which then generates candidate main arguments one at a time. An example of the usage is below:

\textit{
[You are a commentary writing expert. Please complete the main argument in the direction of \{\textbf{direction}\} based on the peg: \{\textbf{peg}\} and event detail: \{\textbf{event detail}\}. The main argument should be profound, concise, and strongly related to the peg. Please provide the main argument:]
}

To enhance the model's ability to generate the main arguments, we design the corresponding SFT data. The input for this SFT data includes retrieved event details based on the peg and the direction of the article. 
The label for this SFT data is the main argument derived from the input. This data construction approach not only facilitates the generation of a helpful main argument for diverse article directions but also ensures its consistency with the initial peg, thus guaranteeing both relevance and alignment in the narrative.

To meet the advanced requirements of argument, these generated main arguments will then get a score from the argument ranking model, based on their novelty, and objectivity. Then these candidate main arguments will be ranked based on the scores.






\subsubsection{Supporting Argument Generation}
This step aims to generate supporting arguments that seamlessly align with both the main argument and the event's details. To achieve this, the system synthesizes the main argument, event details, and a predefined argument structure. Available argument structures include parallel, progressive, and contrasting formats, each facilitating a unique commentary structure.
This integration process enables the LLM to produce relevant supporting arguments. The LLM will decide the number of supporting arguments $m$ itself. An example of usage could be:

\textit{
[You are a commentary writing expert. Based on the given main argument \{\textbf{main argument}\} of the commentary and event detail \{\textbf{event detail}\}, generate multiple supporting arguments for the commentary. 
The supporting arguments form \{\textbf{structure}\} structure, refining around the \{\textbf{main argument}\} with multi-level, multi-faceted, and multi-angle perspectives. Please provide the supporting arguments:]
}

The SFT data for this step is constructed to facilitate this process. We utilize inputs comprising event details, main arguments, and argument structures. The labels are the marked corresponding supporting arguments. This data construction approach ensures that the model is fine-tuned to produce supporting arguments that enrich and substantiate the main argument effectively.

\subsubsection{Evidence Generation}
In the Evidence Generation step, the system aims to generate accurate and contextually relevant evidence.
The process begins with accessing reference information of a supporting argument from the evidence database to ensure veracity, effectively mitigating the hallucinations. This reference information, along with the provided main and supporting arguments, serves as the input. Then, the LLM will generate evidence that is tailored to align precisely with the given supporting argument. An example of the usage is below:

\textit{[You are a commentary writing expert. Surrounding the main argument \{\textbf{main argument}\}, please use the evidence provided in the reference information, including dates, data, viewpoints, core content, etc., to continue writing evidence in the commentary to support the supporting argument \{\textbf{supporting argument}\}. Please annotate the corresponding reference information numbers in the continuation. 
Reference information: \{\textbf{reference}\}. 
Please provide the evidence:
]}

The SFT data for evidence generation is structured with inputs including reference information, main and supporting arguments. The output label is the evidence associated with the corresponding supporting argument. These elements guide the model in generating evidence that is precise and contextually relevant to the provided supporting arguments.

\subsubsection{Article Combination}
This step is aimed at generating the title and ending, then forming the overall commentary.
To achieve this, the system integrates the preceding event details, main arguments, supporting arguments, and evidence as inputs. Following this integration, the LLM then outputs the title and ending. An example of the usage for ending generation is below:

\textit{
[You are a commentary writing expert. Please write a conclusion for the article, maintaining smooth language, consistent style, and logical coherence with the preceding text. The preceding text is as follows: \{\textbf{preceding text}\}. Please provide the ending:]
}

The usage for title generation is similar to the ending generation.

To facilitate the model in generating context-appropriate and coherent title and ending, the SFT data is constructed with inputs including the event details, main argument, combined supporting arguments and evidence. The label for this data is the corresponding title and ending. This structured approach ensures that the model is adept at crafting titles and endings that effectively encapsulate the various dimensions of the commentary, providing a fitting start and end to the narrative.

After generating the title and ending, the system will combine all the output to form a complete commentary.

\subsection{Auxiliary Components}
In this section, we introduce two auxiliary components that assist Xinyu in meeting advanced requirements: the argument ranking model and the evidence database.

\subsubsection{Argument Ranking Model}
The argument ranking model plays a crucial role in the main argument generation process by assessing and ranking candidate main arguments. 
This aids users in selecting the most compelling argument.


Developing the ranking model presents a central challenge due to the subjective nature of assessing arguments, which lack universally accepted standards, unlike quantifiable metrics. 
For example, evaluating the generated arguments based on factors like novelty is cumbersome.

To address this challenge, we train a BERT-based scoring model with a pairwise loss function. 
This approach converts the ranking challenge into a series of binary comparisons, simplifying the task to discerning relative superiority between pairs of arguments. 
The loss function is defined as: 
\begin{equation}
    \mathcal{L}(x) = \sum \Phi(f(x_a) - f(x_b))
\end{equation}
where $f(x)$ represents the scoring function for a given argument $x$, and $\Phi$ is a non-linear transformation applied to the calculated difference between the two arguments $x_a$ and $x_b$. 


The quantity of \textit{likes} on articles from opinion-sharing platforms, such as Zhihu, is often indicative of the novelty and objectivity of the arguments they present. Consequently, this metric is leveraged to assess the quality of the arguments within these articles. In the process of constructing the training dataset, we collect articles from such platforms, utilizing the number of likes as a criterion to establish a partial order among pairs of texts, and the dataset consists of 240,000 text pairs. This order serves to reflect their relative quality. In the inference phase, we translate the order to numerical score, and the scores of the candidate main arguments are utilized to rank these arguments.

\subsubsection{Evidence Database Construction} 

In pursuit of generating convincing evidence, we construct an Evidence Database to store Chinese knowledge sourced from events and books for retrieval.

For the events knowledge, we first legally collect the daily updated article titles on the website's hot list, and then prompt the LLM to complete the following four tasks given the article title:
(1) summarize the event related to the article title;
(2) determine which direction (e.g., technology, finance) the event belongs to; 
(3) extract the six elements of the event, including time, location, person, cause, process and result; 
(4) describe the event in a paragraph based on the six elements. 

For knowledge from books, 
we gather classic works in law, finance, and various other subjects legally, segmenting the contents of these books into chunks and storing them within the database.

The evidence database is built upon 200,000 event knowledge data and 110,000 book knowledge data.
Following the construction phase, we implement the ElasticSearch engine, anchored to the evidence database, to enhance retrieval capabilities. During retrieval, the supporting argument is inputted, prompting the fetching of the 
$k$ most pertinent references from the database. These references are then fed into the Large Language Model (LLM) to generate evidence in support of the arguments. In practical application, the value of $k$ is determined by the similarity score between the input argument and the existing knowledge, and we set a predefined threshold at 0.6 in the experiment.
In addition, to maintain access to the most current event knowledge, we continuously collect data from online platforms and update our database daily.


%% file: 5_Experiment.tex
\section{Experiment}

\subsection{Evaluation Metrics of Commentary}

\paragraph{Automatic evaluation}
Existing automatic generation evaluation metrics, including but not limited to ROUGE and BLEU\citep{papineni2002bleu}, mainly focus on the degree of similarity to a reference text. However, in the context of commentary generation, the inherent diversity of the commentary content poses a significant challenge to these similarity-based metrics, often leading to an incomplete evaluation. To address this limitation, we propose a novel evaluation metric that assesses commentaries across five distinct dimensions: 
$\bullet$ \textbf{Structure Soundness}: clarity of the hierarchy, compactness of the writing, and rationality of the layout; 
$\bullet$ \textbf{Logic Consistency}: consistency of the content, rationality of the argument, and thoughtfulness; 
$\bullet$ \textbf{Argument Quality}: freshness and directionality of the topic conception;
$\bullet$ \textbf{Evidence Support}: specificity and appropriateness of the evidence used; and 
$\bullet$ \textbf{Language Finesse}: fluency, depth, and vividness of the expression style. Besides, we calculate the average of the five scores as \textbf{Overall}.

The prompt templates are as follows:

\textit{[You are an expert in scoring generated commentaries. Please rate your answers from the \{\textbf{perspective}\} perspective based on the provided commentary. The scoring criteria are:}

\textit{
(1) 10 points represent...
(2) 8 points represent...
(3) 6 points ...
}

\textit{Please output a line that contains only one value representing the score. Please avoid any potential biases, and ensure that there are no factors other than the text that affect your judgment.]}

These dimensions are chosen to encompass both the fundamental and advanced requirements of commentary. Structural soundness and logical consistency constitute the fundamental requirements, ensuring a well-organized and logically coherent commentary. Conversely, the quality of argumentation and the adequacy of evidentiary support represent the advanced requirements, reflecting the depth and persuasiveness of the commentary. The dimensions are scored on a scale of 1-10, with 1 being the lowest and 10 the highest.

In our experiments, we utilize GPT-4 for automatic evaluation by crafting specific prompts for each dimension. To validate GPT-4's accuracy, we compare its scores against those from human annotators for 30 randomly selected commentaries, calculating the Pearson correlation coefficient \citep{cohen2009pearson} for each dimension. As Tab. \ref{metric} illustrates, the Pearson correlation coefficient of each dimension surpasses 0.6, which proves GPT-4 is competent for this task.

\paragraph{Human evaluation}
In our ablation study, we assess the \textbf{Timeliness} of evidence. Due to the training limitations of GPT-4, which is based on data available only up to a specific date, it is not equipped to accurately ascertain the recency of evidence. Therefore, this aspect is evaluated through human judgment. The scoring for this metric ranges from 1 to 10, where 1 represents the lowest and 10 the highest possible score.


\begin{table}
\begin{tabular}{@{}lccccc@{}}
\toprule
Dimension   & Structure & Logic & Argu. & Evidence & Language \\ \midrule
Pearson's r & 0.66 & 0.69 & 0.73  & 0.66 & 0.64 \\ \bottomrule
\end{tabular}
\caption{Consistency analysis of Human and GPT-4 on five dimensions. Arug. refers to Argument, and r means the Pearson correlation coefficient.}
\label{metric}
\end{table}

\subsection{Experiment Settings}
\begin{figure}[t]
\centering
\includegraphics[width=0.35\textwidth]{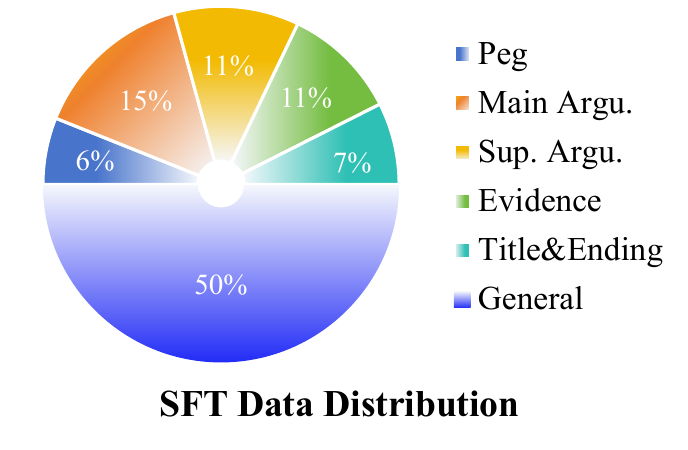}
\caption{
The distribution of SFT datast. Argu. refers to Argument, Sup. means Supporting.
}
\label{fig:sft}
\end{figure}

\paragraph{Implementation} The base model of Xinyu is LLaMA2-13B \citep{llama}, and we specifically adapted it to better accommodate the nuances of the Chinese language. This adaptation involved expanding the LLaMA-13B tokenizer with an additional 28,000 Chinese tokens. To further optimize the model, we continued pre-training on LLaMA-13B using a corpus comprising 500B tokens, which contains both English and Chinese corpus. For supervised fine-tuning, we not only utilized the dataset introduced in Section 4 but also incorporated the general SFT dataset\footnote{\url{https://huggingface.co/datasets/BelleGroup/train_2M_CN}} to maintain consistency with the data distribution of previous training phases. The amount of SFT data is 400,000 and the distribution of it is shown in Figure. \ref{fig:sft}. 
Our training process leveraged the Megatron-DeepSpeed framework. The continued pre-training phase lasted 20 days on 128 Nvidia A800 80G GPUs, while the supervised fine-tuning (SFT) process took 2 days on 8 Nvidia A800 80G GPUs.

\begin{table*}
\begin{tabular}{lcccccc}
\hline
\textbf{Methods} & \textbf{Overall} & \textbf{Structure} & \textbf{Logic} & \textbf{Argument} & \textbf{Evidence} & \textbf{Language} \\ \hline
Baichuan2-Turbo \citep{baichuan} & 7.65 & 8.11* & 8.28 & 8.06 & 5.63 & 8.17 \\
Qwen-72B \citep{qwen} & 7.37 & 7.90 & 7.88 & 7.87 & 5.28 & 7.90 \\
GLM-4 \citep{glm} & 7.72 & 8.11* & 8.35 & 8.08* & 5.82 & 8.24 \\
ERNIE-4 \citep{ernie} & 7.71 & 8.05 & 8.35 & 8.03 & 5.73 & 8.38* \\
GPT-3.5-Turbo \citep{instructgpt} & 7.70 & 8.00 & 8.08 & 8.00 & 6.39 & 8.05 \\
GPT-4 \citep{gpt4} & 7.78 & 8.05 & 8.40* & 8.05 & 6.02 & 8.38* \\\hline
Baichuan2-13B-Chat \cite{baichuan} & 7.37 & \textbf{8.11}* & 8.00 & 7.92 & 4.78 & \textbf{8.05} \\
Qwen-14B-Chat \citep{qwen} & 7.25 & 8.05 & 7.88 & 7.85 & 4.55 & 7.95 \\
InternLM-20B-Chat \citep{internlm} & 7.26 & 7.80 & 7.83 & 7.88 & 4.83 & 8.00 \\
Xinyu (based on fine-tuned LLaMA2-13B) & \textbf{7.93}* & 8.00 & \textbf{8.20} & \textbf{8.00} & \textbf{7.41}* & \textbf{8.05} \\ \hline
\end{tabular}
\caption{Evaluation of commentary generated by baseline LLMs using GPT-4. Reference is the published commentaries. `Bold' indicates the highest score within the 20B scale baselines, and an asterisk (*) denotes the highest score among all baselines.}
\label{tab:main}
\end{table*}

\begin{table*}
\begin{tabular}{lcccccc}
\hline
\textbf{Base Model} & \textbf{Overall} & \textbf{Structure} & \textbf{Logic} & \textbf{Argument} & \textbf{Evidence} & \textbf{Language} \\ \hline
Qwen-72B-Chat \citep{qwen} & 7.82 & 7.80 & 7.85 & 7.75 & 7.80 & 7.90 \\
Baichuan2-Turbo \citep{baichuan} & 7.91 & 8.00 & 7.93 & 7.70 & 8.11 & 7.80 \\
GPT-4 \citep{gpt4} & 8.30* & 8.10* & 8.58* & 8.15* & 8.17* & 8.50* \\ \hline
Baichuan2-13B \citep{baichuan} & 6.31 & 5.50 & 6.55 & 6.78 & 6.49 & 6.23 \\
Qwen-14B-Chat \citep{qwen} & 6.22 & 5.60 & 6.25 & 6.25 & 6.80 & 6.20 \\
Xinyu & \textbf{7.93} & \textbf{8.00} & \textbf{8.20} & \textbf{8.00} & \textbf{7.41} & \textbf{8.05} \\ \hline
\end{tabular}
\caption{GPT-4's evaluation of commentary generated by our framework with different base models. `Bold' indicates the highest score within the 20B scale baselines, and an asterisk (*) denotes the highest score among all baselines.}
\label{tab:frame}
\end{table*}

\begin{table*}
\begin{tabular}{lcccccc}
\hline
\textbf{Methods} & \textbf{Overall} & \textbf{Structure} & \textbf{Logic} & \textbf{Argument} & \textbf{Evidence} & \textbf{Language} \\ \hline
 w/o Ranking & 7.85 & 7.93 & 8.20 & 7.80 & 7.31 & 8.02 \\
Xinyu & \textbf{7.93} & \textbf{8.00} & \textbf{8.20} & \textbf{8.00} & \textbf{7.41} & \textbf{8.05} \\
\hline
\end{tabular}
\caption{Results of ablation study on Argument Ranking Model.}
\label{tab:ranking}
\end{table*}

\paragraph{Baselines} 
We employ the following methods as our baselines:

Baichuan2 \citep{baichuan} represents a series of large-scale, multilingual language models trained from scratch on 2.6 trillion tokens. We select \textbf{Baichuan2-13B-Chat} and \textbf{Baichuan2-Turbo} as our baseline models. Qwen \citep{qwen} is a comprehensive series of language models featuring a range of models with varying parameter counts. In this context, we choose \textbf{Qwen-72B} and \textbf{Qwen-72B-Chat} as our baseline models. InternLM \citep{internlm} consists of a series of multilingual foundation models and chat models, with \textbf{InternLM-20B-Chat} selected as the baseline model. GLM \citep{glm} is a series of bilingual (English and Chinese) pre-trained language models, for which we use \textbf{GLM-4} as the baseline model. ERNIE \citep{ernie} serves as a unified framework for pre-training large-scale knowledge-enhanced models, with \textbf{ERNIE-4} chosen as our baseline model. Finally, GPT is a series of large language models released by OpenAI, with \textbf{GPT-3.5-Turbo} \citep{instructgpt} and \textbf{GPT-4} \citep{gpt4} used as baseline models.

\paragraph{Test Cases} 
For our test cases, we have carefully chosen 41 commentaries from the ``Three Commentaries'' section of People's Daily Online\footnote{\url{http://opinion.people.com.cn/GB/8213/420650/index.html}}. This selection encompasses a diverse range of topics including economics, livelihood, technology, culture, social issues, sports, and art, reflecting current news and societal trends. The prominence of the site and the authoritative nature of the series ensure that these articles represent high-quality journalistic commentary. In our baselines, we employ a one-step generation process using Event Detail, and extra Title, Argument, and Evidence, which are from the real commentary of the event. This setup is designed to emulate real-world scenarios where commentators use LLMs. The translated prompt in English is:

\textit{[I will provide you with a news background: \{Event detail\}}

\textit{Based on this news, with the title '{Title}', please create a commentary article. The article should have clear and profound arguments, true and abundant evidence, smooth logical reasoning, reasonable structure, and appropriate commentary language. Your article should reference the following argument and evidence:}
\textit{
Argument 1: \{Argument 1\}}
\textit{
Evidence 1: \{Evidence 1\}}
\textit{
Argument 2: \{Argument 2\}}
\textit{
Evidence 2: \{Evidence 2\}}
\textit{
Argument 3: \{Argument 3\}}
\textit{
Evidence 3: \{Evidence 3\}]}

The number of \textit{Arguments} and \textit{Evidence} will be 0 to 3.

However, for Xinyu, we only supply the event details to evaluate the effectiveness of our approach in generating commentary.

\paragraph{Ablation}
In our ablation experiments, we adopt the following configurations to assess specific components:

1) Evaluating the impact of the framework: While maintaining the overall framework intact, we replace the Xinyu-13B model with alternative large language models (LLMs) to determine the effectiveness of the framework.

2) Assessing ranking efficiency: We eliminate the argument ranking component, allowing the LLM to directly generate a main argument without a predefined direction, to evaluate the ranking model's contribution.

3) Investigating the role of RAG: By omitting the retrieval process from the evidence database, we let the LLM independently produce the evidence, aiming to understand the significance of the RAG component in enhancing model performance and the effectiveness of our construed evidence database. Moreover, we assess the timeliness of the generated evidence especially.

4) Measuring the practicality of Xinyu: We compared the quality of commentary articles generated by human experts with a journalism background with the aid of Xinyu, against those written purely manually, while also comparing the time spent. We randomly select 10 test cases.




\begin{figure*}[ht]
\centering
\includegraphics[width=0.9\textwidth]
{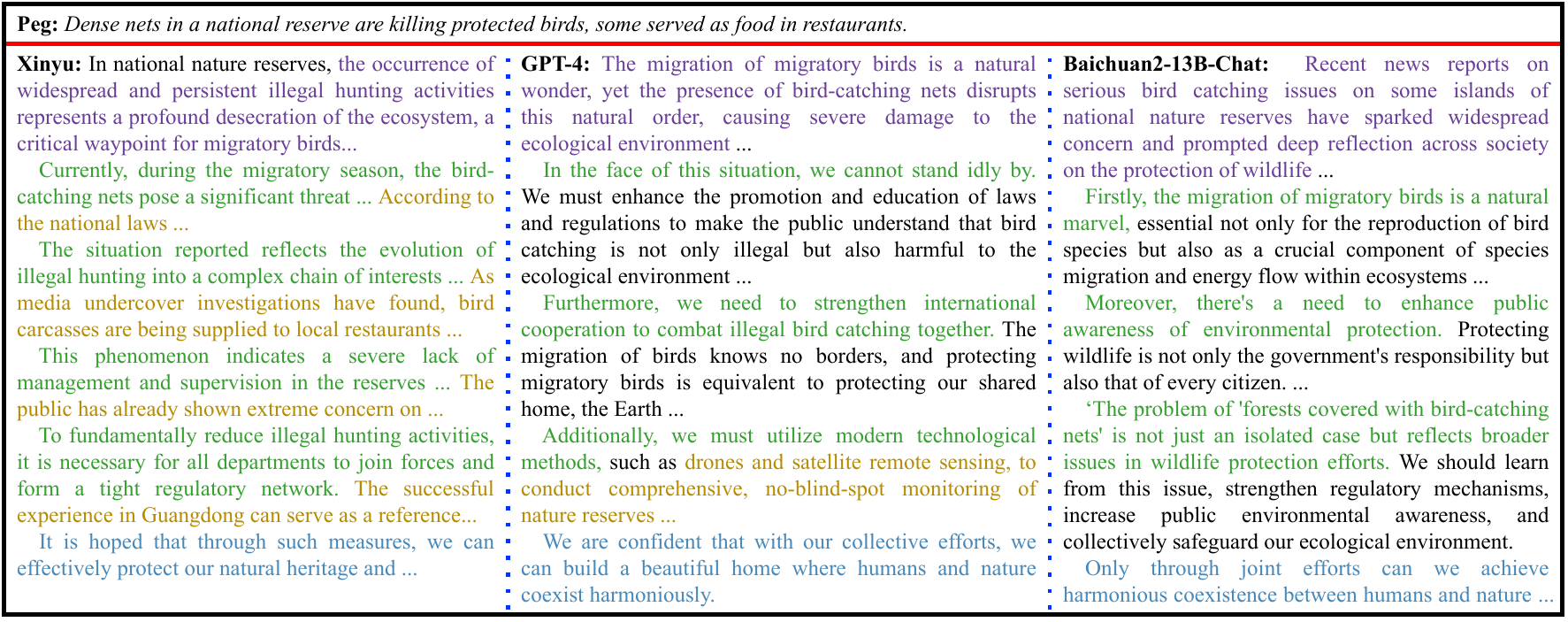}
\caption{Case study. The content is translated from Chinese.}
\label{fig:casestudy}
\end{figure*}

\begin{table}
\begin{tabular}{lcc}
\hline
\textbf{Methods} & \textbf{Evidence} & \textbf{Timeliness} \\ \hline
w/o RAG & 5.60 & 8.85 \\
w/ Event & 7.14 & 9.10 \\
w/ Event + Book & \textbf{7.41} & \textbf{9.30} \\ \hline
\end{tabular}
\caption{Results of ablation study on Evidence Database. The Kappa value of Timeliness exceeds 0.78.}
\label{tab:evi}
\end{table}

\begin{figure}[t]
\centering
\includegraphics[width=0.45\textwidth]{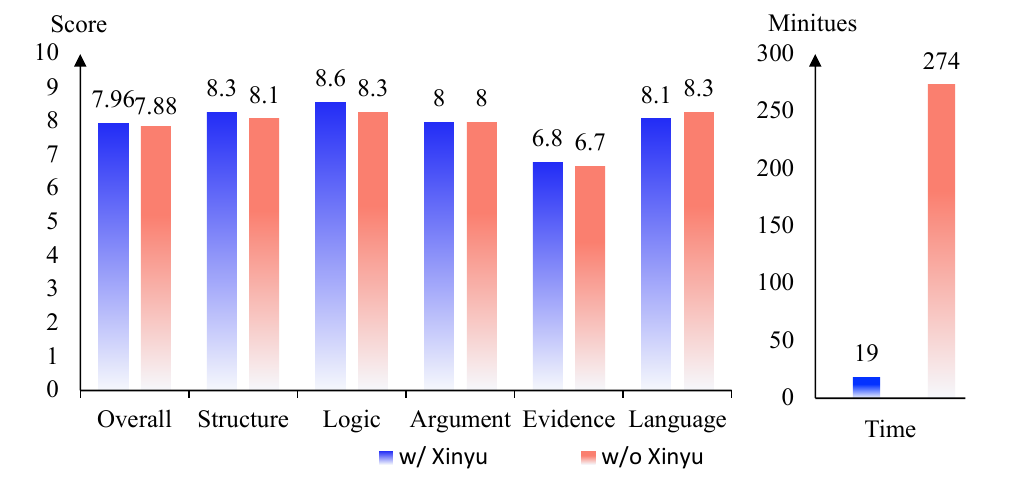}
\caption{
Human vs. Xinyu-Assisted.
}
\label{fig:human}
\end{figure}

\subsection{Experimental Results and Analysis}
Based on the model's size, we split them into two types: LLMs larger than 20B and LLMs smaller than 20B. Tab. \ref{tab:main} shows the results of commentary generation with GPT-4's evaluation. We report the results of the ablation study in Tab. \ref{tab:frame}, Tab. \ref{tab:ranking}, Tab. \ref{tab:evi}, and Figure. \ref{fig:human}.

\paragraph{Results of commentary generation} 

From Tab. \ref{tab:main}, we can conclude that:
(1) Generally, the bigger the language model's size, the better it does. However, GPT-4's leading advantage in this task is not as pronounced as in other generative tasks.
(2) When looking at models within the 20 billion parameter size, our method achieved the best results in most of the metrics.
(3) Compared to large-scale LLMs such as GPT-4, our method attained the best overall score, primarily due to our superior performance in the advanced requirements of argument and evidence.
(4) There's not a huge gap between the scores of the different methods. This is largely because GPT-4 is generally not harsh in its scoring, rarely giving out very low scores.

\paragraph{Result of ablation study} From Tab. \ref{tab:frame}, we have the following observations:
(1) Our framework significantly enhances the performance of large-sized base models. For instance, the overall score of Qwen-72B increased from 7.37 to 7.82. 
(2) GPT-4 achieved the best performance with an overall score of 8.3, and our Xinyu ranks just behind GPT-4. Considering the size of the model, our method has greater potential in practice.
(3) For the 20B scale base models such as Qwen-14B-Chat, using our framework actually decreased their performance. This might be due to these base models' inherent limitations in generating text step-by-step. This also demonstrates the effectiveness of our SFT.

From Tab. \ref{tab:ranking}, we have the following observations:
(1) The implementation of the argument ranking model has significantly improved the effectiveness of argumentation, underscoring its impact. 
(2) Enhancements are observed across all metrics, illustrating the interrelationship among these aspects of commentary.

From Tab. \ref{tab:evi}, we can find that:
(1) The implementation of retrieval augmented generation (RAG) significantly enhances the generation of evidence, with scores improving from 5.60 to 7.41. Additionally, the timeliness of the generated evidence also saw an increase, rising from 8.85 to 9.30.
(2) After incorporating the book dataset into RAG, its performance experienced further improvements.
(3) The improvement proves the effectiveness of our evidence dataset.

From Figure. \ref{fig:human}, we have the following observations:
(1) Utilizing Xinyu's assistance can significantly increase writing speed, and the average time for a commentary speeds up from more than 4 hours to 20 mins.
(2) Moreover, commentaries generated with LLMs have achieved the same scores as manual writing, demonstrating the practicality of our system.

\paragraph{Case study} 
Figure \ref{fig:casestudy} presents three commentaries on a certain peg generated by Xinyu, GPT-4, and Baichuan2-13B-Chat, respectively. All three commentaries exhibit good language fluency and structural coherence, highlighting the capabilities of these large language models (LLMs). However, the commentary from Baichuan2-13B-Chat focuses solely on facts without offering specific arguments. In contrast, both GPT-4 and Xinyu provide detailed arguments. Notably, Xinyu's commentary stands out by presenting more convincing evidence and demonstrating a logical correlation in its supporting arguments.


%% file: 6_Conclusion.tex
\section{Conclusion}
In this paper, we introduce Xinyu, an innovative commentary generation system based on large language models (LLMs) designed to enhance the efficiency of commentators. Our approach involves breaking down the generation process into five steps, with supervised fine-tuning (SFT) applied to each step to ensure the output is well-structured and coherent, addressing the basic requirements of commentary. To fulfill the higher demands for novelty and persuasiveness, we develop an argument ranking model and employ retrieval-augmented generation (RAG) techniques for evidence generation. For RAG, we have compiled an evidence database comprising both current events and classical books. To better measure the generated commentaries, we design a comprehensive evaluation method with 5 distinct perspectives. Our comprehensive experiments demonstrate the system's effectiveness. Remarkably, in practical applications, Xinyu has reduced the average commentary creation time from 4 hours to just 20 minutes and maintained the quality.

In the future, we will consider the following directions to enhance our system: 1) Improve evidence recall accuracy, ensuring relevance to the arguments; 2) Utilize Reinforcement Learning with Human Feedback (RLHF) to better align commentaries with human preferences and specific writing styles.

%% file: Acknowledgment.tex
\section*{Acknowledgments}
This work was supported by the National Natural Science Foundation of China (62441605, 62376243, 62037001, U20A20387), and the Starry Night Science Fund of Zhejiang University Shanghai Institute for Advanced Study (SN-ZJU-SIAS-0010).

Finally, we would like to thank the anonymous reviewers for their helpful feedback and suggestions.

%% file: 7_Appendix.tex
\appendix
\section{appendix}


\subsection{Overall Generation Process}

This section presents a complete example corresponding to each step in section 4.1. 
The overall process consists of Peg Generation -> Main Argument Generation -> Supporting Argument Generation -> Evidence Generation -> Ending Generation \& Title Generation. 
All the content is translated from Chinese.

\begin{figure*}[t]
\centering
\includegraphics[width=1\textwidth]{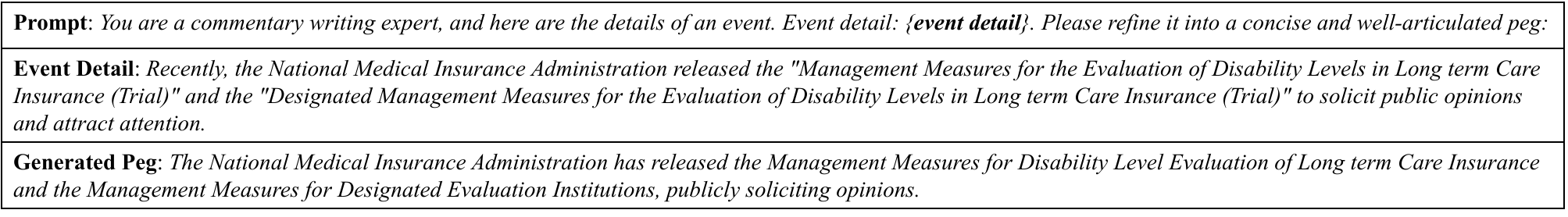}
\vspace{-20pt}
\caption{Peg Generation.}
\label{peg}
\end{figure*}

\begin{figure*}[t]
\centering
\includegraphics[width=1\textwidth]{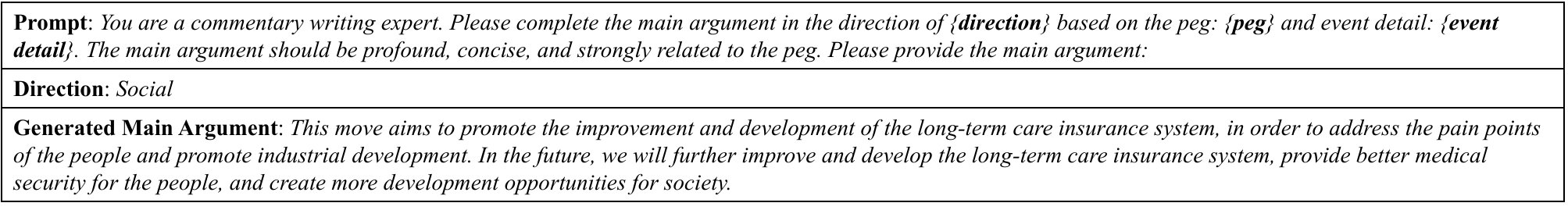}
\vspace{-20pt}
\caption{Main Argument Generation.}
\label{main}
\end{figure*}

\begin{figure*}[t]
\centering
\includegraphics[width=1\textwidth]{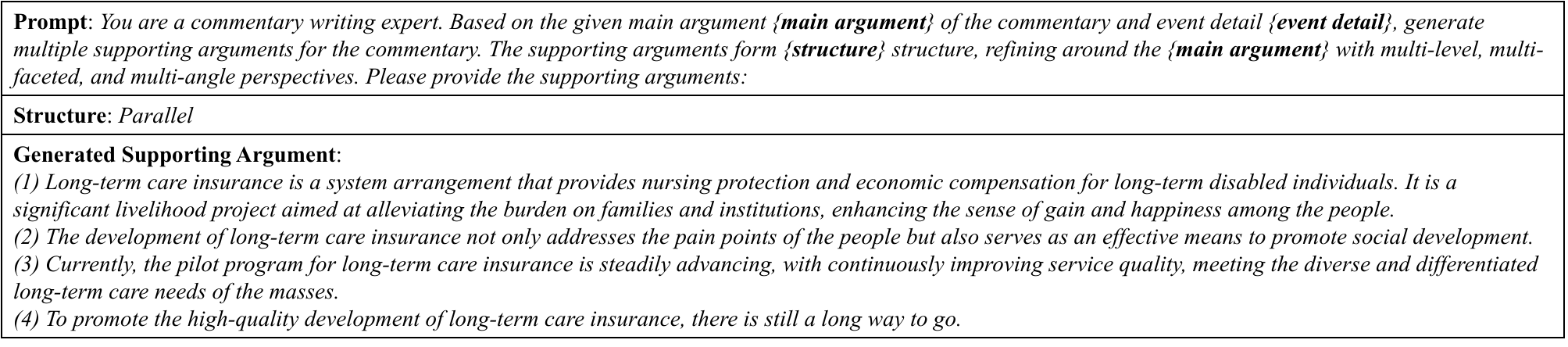}
\vspace{-20pt}
\caption{Supporting Argument Generation.}
\label{sup}
\end{figure*}

\begin{figure*}[t]
\centering
\includegraphics[width=1\textwidth]{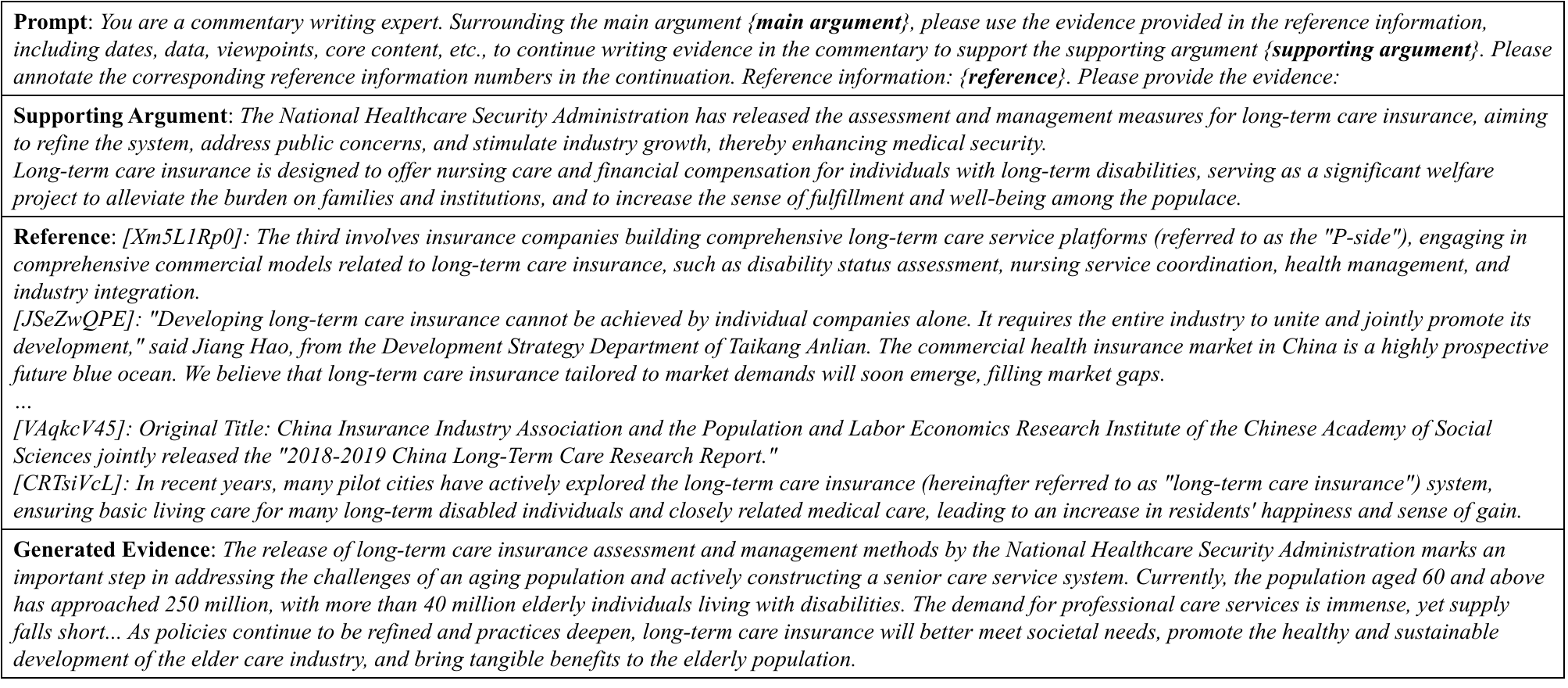}
\caption{Evidence Generation.}
\label{evidence}
\end{figure*}

\begin{figure*}[t]
\centering
\includegraphics[width=1\textwidth]{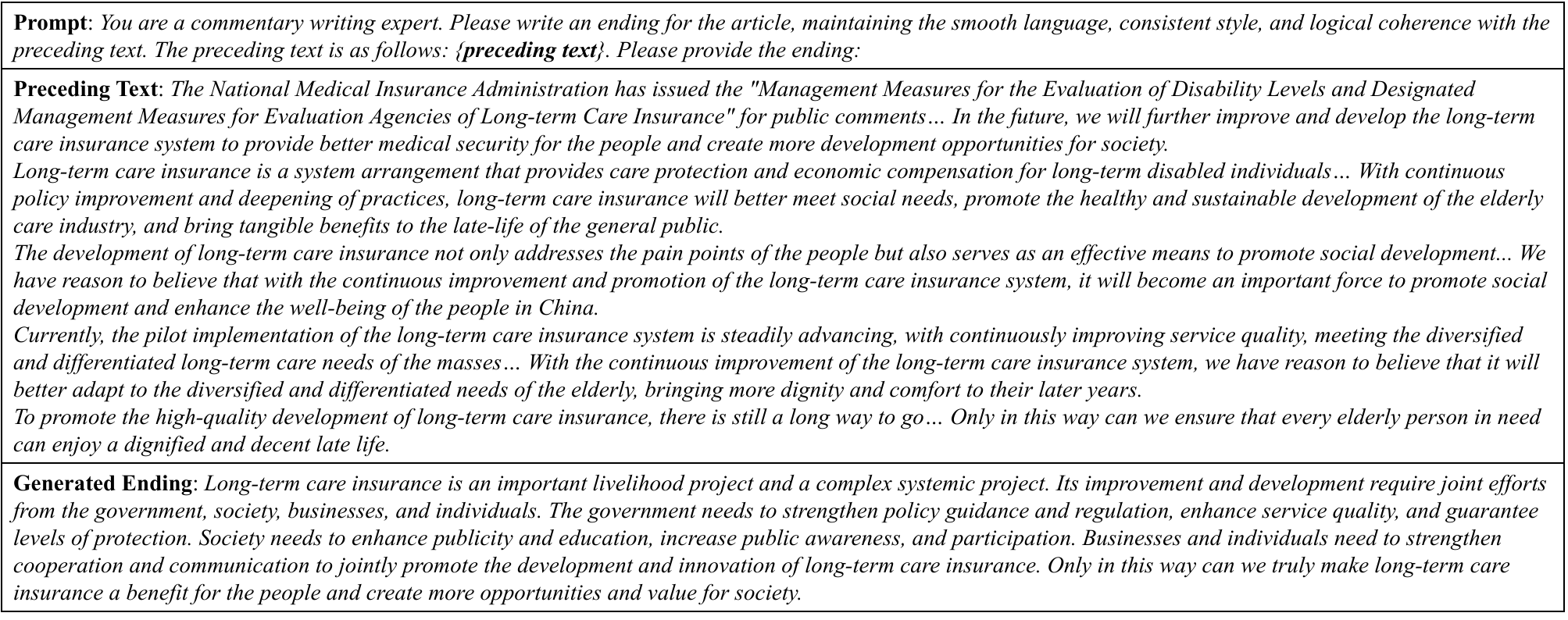}
\vspace{-20pt}
\caption{Ending Generation.}
\label{ending}
\end{figure*}

\begin{figure*}[t]
\centering
\includegraphics[width=1\textwidth]{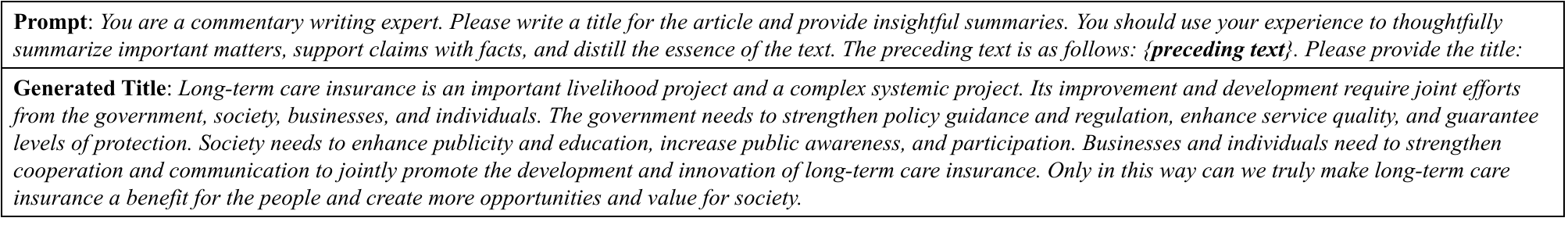}
\vspace{-20pt}
\caption{Title Generation.}
\label{title}
\end{figure*}

%% file: Main.bbl

\begin{thebibliography}{42}


\ifx \showCODEN    \undefined \def \showCODEN     #1{\unskip}     \fi
\ifx \showDOI      \undefined \def \showDOI       #1{#1}\fi
\ifx \showISBNx    \undefined \def \showISBNx     #1{\unskip}     \fi
\ifx \showISBNxiii \undefined \def \showISBNxiii  #1{\unskip}     \fi
\ifx \showISSN     \undefined \def \showISSN      #1{\unskip}     \fi
\ifx \showLCCN     \undefined \def \showLCCN      #1{\unskip}     \fi
\ifx \shownote     \undefined \def \shownote      #1{#1}          \fi
\ifx \showarticletitle \undefined \def \showarticletitle #1{#1}   \fi
\ifx \showURL      \undefined \def \showURL       {\relax}        \fi
\providecommand\bibfield[2]{#2}
\providecommand\bibinfo[2]{#2}
\providecommand\natexlab[1]{#1}
\providecommand\showeprint[2][]{arXiv:#2}

\bibitem[Adeshola and Adepoju(2023)]%
        {adeshola2023opportunities}
\bibfield{author}{\bibinfo{person}{Ibrahim Adeshola} {and} \bibinfo{person}{Adeola~Praise Adepoju}.} \bibinfo{year}{2023}\natexlab{}.
\newblock \showarticletitle{The opportunities and challenges of ChatGPT in education}.
\newblock \bibinfo{journal}{\emph{Interactive Learning Environments}} (\bibinfo{year}{2023}), \bibinfo{pages}{1--14}.
\newblock


\bibitem[Bai et~al\mbox{.}(2023)]%
        {qwen}
\bibfield{author}{\bibinfo{person}{Jinze Bai}, \bibinfo{person}{Shuai Bai}, \bibinfo{person}{Yunfei Chu}, \bibinfo{person}{Zeyu Cui}, \bibinfo{person}{Kai Dang}, \bibinfo{person}{Xiaodong Deng}, \bibinfo{person}{Yang Fan}, \bibinfo{person}{Wenbin Ge}, \bibinfo{person}{Yu Han}, \bibinfo{person}{Fei Huang}, \bibinfo{person}{Binyuan Hui}, \bibinfo{person}{Luo Ji}, \bibinfo{person}{Mei Li}, \bibinfo{person}{Junyang Lin}, \bibinfo{person}{Runji Lin}, \bibinfo{person}{Dayiheng Liu}, \bibinfo{person}{Gao Liu}, \bibinfo{person}{Chengqiang Lu}, \bibinfo{person}{Keming Lu}, \bibinfo{person}{Jianxin Ma}, \bibinfo{person}{Rui Men}, \bibinfo{person}{Xingzhang Ren}, \bibinfo{person}{Xuancheng Ren}, \bibinfo{person}{Chuanqi Tan}, \bibinfo{person}{Sinan Tan}, \bibinfo{person}{Jianhong Tu}, \bibinfo{person}{Peng Wang}, \bibinfo{person}{Shijie Wang}, \bibinfo{person}{Wei Wang}, \bibinfo{person}{Shengguang Wu}, \bibinfo{person}{Benfeng Xu}, \bibinfo{person}{Jin Xu}, \bibinfo{person}{An Yang}, \bibinfo{person}{Hao Yang},
  \bibinfo{person}{Jian Yang}, \bibinfo{person}{Shusheng Yang}, \bibinfo{person}{Yang Yao}, \bibinfo{person}{Bowen Yu}, \bibinfo{person}{Hongyi Yuan}, \bibinfo{person}{Zheng Yuan}, \bibinfo{person}{Jianwei Zhang}, \bibinfo{person}{Xingxuan Zhang}, \bibinfo{person}{Yichang Zhang}, \bibinfo{person}{Zhenru Zhang}, \bibinfo{person}{Chang Zhou}, \bibinfo{person}{Jingren Zhou}, \bibinfo{person}{Xiaohuan Zhou}, {and} \bibinfo{person}{Tianhang Zhu}.} \bibinfo{year}{2023}\natexlab{}.
\newblock \showarticletitle{Qwen Technical Report}.
\newblock \bibinfo{journal}{\emph{CoRR}}  \bibinfo{volume}{abs/2309.16609} (\bibinfo{year}{2023}).
\newblock
\urldef\tempurl%
\url{https://doi.org/10.48550/ARXIV.2309.16609}
\showDOI{\tempurl}
\showeprint[arXiv]{2309.16609}


\bibitem[Cardenas et~al\mbox{.}(2023)]%
        {science}
\bibfield{author}{\bibinfo{person}{Ronald Cardenas}, \bibinfo{person}{Bingsheng Yao}, \bibinfo{person}{Dakuo Wang}, {and} \bibinfo{person}{Yufang Hou}.} \bibinfo{year}{2023}\natexlab{}.
\newblock \showarticletitle{'Don't Get Too Technical with Me': {A} Discourse Structure-Based Framework for Automatic Science Journalism}. In \bibinfo{booktitle}{\emph{Proceedings of the 2023 Conference on Empirical Methods in Natural Language Processing, {EMNLP} 2023, Singapore, December 6-10, 2023}}, \bibfield{editor}{\bibinfo{person}{Houda Bouamor}, \bibinfo{person}{Juan Pino}, {and} \bibinfo{person}{Kalika Bali}} (Eds.). \bibinfo{publisher}{Association for Computational Linguistics}, \bibinfo{pages}{1186--1202}.
\newblock
\urldef\tempurl%
\url{https://aclanthology.org/2023.emnlp-main.76}
\showURL{%
\tempurl}


\bibitem[Cheng et~al\mbox{.}(2023)]%
        {cheng2023now}
\bibfield{author}{\bibinfo{person}{Szu-Wei Cheng}, \bibinfo{person}{Chung-Wen Chang}, \bibinfo{person}{Wan-Jung Chang}, \bibinfo{person}{Hao-Wei Wang}, \bibinfo{person}{Chih-Sung Liang}, \bibinfo{person}{Taishiro Kishimoto}, \bibinfo{person}{Jane Pei-Chen Chang}, \bibinfo{person}{John~S Kuo}, {and} \bibinfo{person}{Kuan-Pin Su}.} \bibinfo{year}{2023}\natexlab{}.
\newblock \showarticletitle{The now and future of ChatGPT and GPT in psychiatry}.
\newblock \bibinfo{journal}{\emph{Psychiatry and clinical neurosciences}} \bibinfo{volume}{77}, \bibinfo{number}{11} (\bibinfo{year}{2023}), \bibinfo{pages}{592--596}.
\newblock


\bibitem[Cohen et~al\mbox{.}(2009)]%
        {cohen2009pearson}
\bibfield{author}{\bibinfo{person}{Israel Cohen}, \bibinfo{person}{Yiteng Huang}, \bibinfo{person}{Jingdong Chen}, \bibinfo{person}{Jacob Benesty}, \bibinfo{person}{Jacob Benesty}, \bibinfo{person}{Jingdong Chen}, \bibinfo{person}{Yiteng Huang}, {and} \bibinfo{person}{Israel Cohen}.} \bibinfo{year}{2009}\natexlab{}.
\newblock \showarticletitle{Pearson correlation coefficient}.
\newblock \bibinfo{journal}{\emph{Noise reduction in speech processing}} (\bibinfo{year}{2009}), \bibinfo{pages}{1--4}.
\newblock


\bibitem[Cui et~al\mbox{.}(2023)]%
        {chatlaw}
\bibfield{author}{\bibinfo{person}{Jiaxi Cui}, \bibinfo{person}{Zongjian Li}, \bibinfo{person}{Yang Yan}, \bibinfo{person}{Bohua Chen}, {and} \bibinfo{person}{Li Yuan}.} \bibinfo{year}{2023}\natexlab{}.
\newblock \showarticletitle{ChatLaw: Open-Source Legal Large Language Model with Integrated External Knowledge Bases}.
\newblock \bibinfo{journal}{\emph{CoRR}}  \bibinfo{volume}{abs/2306.16092} (\bibinfo{year}{2023}).
\newblock
\urldef\tempurl%
\url{https://doi.org/10.48550/ARXIV.2306.16092}
\showDOI{\tempurl}
\showeprint[arXiv]{2306.16092}


\bibitem[Gan et~al\mbox{.}(2023)]%
        {edu-llm}
\bibfield{author}{\bibinfo{person}{Wensheng Gan}, \bibinfo{person}{Zhenlian Qi}, \bibinfo{person}{Jiayang Wu}, {and} \bibinfo{person}{Jerry~Chun{-}Wei Lin}.} \bibinfo{year}{2023}\natexlab{}.
\newblock \showarticletitle{Large Language Models in Education: Vision and Opportunities}.
\newblock \bibinfo{journal}{\emph{CoRR}}  \bibinfo{volume}{abs/2311.13160} (\bibinfo{year}{2023}).
\newblock
\urldef\tempurl%
\url{https://doi.org/10.48550/ARXIV.2311.13160}
\showDOI{\tempurl}
\showeprint[arXiv]{2311.13160}


\bibitem[Gao et~al\mbox{.}(2022)]%
        {hyde}
\bibfield{author}{\bibinfo{person}{Luyu Gao}, \bibinfo{person}{Xueguang Ma}, \bibinfo{person}{Jimmy Lin}, {and} \bibinfo{person}{Jamie Callan}.} \bibinfo{year}{2022}\natexlab{}.
\newblock \bibinfo{title}{Precise Zero-Shot Dense Retrieval without Relevance Labels}.
\newblock
\newblock
\showeprint[arxiv]{2212.10496}~[cs.IR]


\bibitem[Gao et~al\mbox{.}(2023)]%
        {rag-survey}
\bibfield{author}{\bibinfo{person}{Yunfan Gao}, \bibinfo{person}{Yun Xiong}, \bibinfo{person}{Xinyu Gao}, \bibinfo{person}{Kangxiang Jia}, \bibinfo{person}{Jinliu Pan}, \bibinfo{person}{Yuxi Bi}, \bibinfo{person}{Yi Dai}, \bibinfo{person}{Jiawei Sun}, \bibinfo{person}{Qianyu Guo}, \bibinfo{person}{Meng Wang}, {and} \bibinfo{person}{Haofen Wang}.} \bibinfo{year}{2023}\natexlab{}.
\newblock \showarticletitle{Retrieval-Augmented Generation for Large Language Models: {A} Survey}.
\newblock \bibinfo{journal}{\emph{CoRR}}  \bibinfo{volume}{abs/2312.10997} (\bibinfo{year}{2023}).
\newblock
\urldef\tempurl%
\url{https://doi.org/10.48550/ARXIV.2312.10997}
\showDOI{\tempurl}
\showeprint[arXiv]{2312.10997}


\bibitem[Izacard et~al\mbox{.}(2022)]%
        {izacard2022atlas}
\bibfield{author}{\bibinfo{person}{Gautier Izacard}, \bibinfo{person}{Patrick Lewis}, \bibinfo{person}{Maria Lomeli}, \bibinfo{person}{Lucas Hosseini}, \bibinfo{person}{Fabio Petroni}, \bibinfo{person}{Timo Schick}, \bibinfo{person}{Jane Dwivedi-Yu}, \bibinfo{person}{Armand Joulin}, \bibinfo{person}{Sebastian Riedel}, {and} \bibinfo{person}{Edouard Grave}.} \bibinfo{year}{2022}\natexlab{}.
\newblock \bibinfo{title}{Atlas: Few-shot Learning with Retrieval Augmented Language Models}.
\newblock
\newblock
\showeprint[arxiv]{2208.03299}~[cs.CL]


\bibitem[Li and Li(2023)]%
        {UAE}
\bibfield{author}{\bibinfo{person}{Xianming Li} {and} \bibinfo{person}{Jing Li}.} \bibinfo{year}{2023}\natexlab{}.
\newblock \showarticletitle{AnglE-optimized Text Embeddings}.
\newblock \bibinfo{journal}{\emph{CoRR}}  \bibinfo{volume}{abs/2309.12871} (\bibinfo{year}{2023}).
\newblock
\urldef\tempurl%
\url{https://doi.org/10.48550/ARXIV.2309.12871}
\showDOI{\tempurl}
\showeprint[arXiv]{2309.12871}


\bibitem[Li et~al\mbox{.}(2023)]%
        {chatdoctor}
\bibfield{author}{\bibinfo{person}{Yunxiang Li}, \bibinfo{person}{Zihan Li}, \bibinfo{person}{Kai Zhang}, \bibinfo{person}{Ruilong Dan}, {and} \bibinfo{person}{You Zhang}.} \bibinfo{year}{2023}\natexlab{}.
\newblock \showarticletitle{ChatDoctor: {A} Medical Chat Model Fine-tuned on LLaMA Model using Medical Domain Knowledge}.
\newblock \bibinfo{journal}{\emph{CoRR}}  \bibinfo{volume}{abs/2303.14070} (\bibinfo{year}{2023}).
\newblock
\urldef\tempurl%
\url{https://doi.org/10.48550/ARXIV.2303.14070}
\showDOI{\tempurl}
\showeprint[arXiv]{2303.14070}


\bibitem[Liu et~al\mbox{.}(2023a)]%
        {chatcounselor}
\bibfield{author}{\bibinfo{person}{June~M. Liu}, \bibinfo{person}{Donghao Li}, \bibinfo{person}{He Cao}, \bibinfo{person}{Tianhe Ren}, \bibinfo{person}{Zeyi Liao}, {and} \bibinfo{person}{Jiamin Wu}.} \bibinfo{year}{2023}\natexlab{a}.
\newblock \showarticletitle{ChatCounselor: {A} Large Language Models for Mental Health Support}.
\newblock \bibinfo{journal}{\emph{CoRR}}  \bibinfo{volume}{abs/2309.15461} (\bibinfo{year}{2023}).
\newblock
\urldef\tempurl%
\url{https://doi.org/10.48550/ARXIV.2309.15461}
\showDOI{\tempurl}
\showeprint[arXiv]{2309.15461}


\bibitem[Liu et~al\mbox{.}(2023b)]%
        {liu2023ml}
\bibfield{author}{\bibinfo{person}{Yifei Liu}, \bibinfo{person}{Yiquan Wu}, \bibinfo{person}{Yating Zhang}, \bibinfo{person}{Changlong Sun}, \bibinfo{person}{Weiming Lu}, \bibinfo{person}{Fei Wu}, {and} \bibinfo{person}{Kun Kuang}.} \bibinfo{year}{2023}\natexlab{b}.
\newblock \showarticletitle{Ml-ljp: Multi-law aware legal judgment prediction}. In \bibinfo{booktitle}{\emph{Proceedings of the 46th International ACM SIGIR Conference on Research and Development in Information Retrieval}}. \bibinfo{pages}{1023--1034}.
\newblock


\bibitem[Ma et~al\mbox{.}(2023)]%
        {retrieve-read}
\bibfield{author}{\bibinfo{person}{Xinbei Ma}, \bibinfo{person}{Yeyun Gong}, \bibinfo{person}{Pengcheng He}, \bibinfo{person}{Hai Zhao}, {and} \bibinfo{person}{Nan Duan}.} \bibinfo{year}{2023}\natexlab{}.
\newblock \showarticletitle{Query Rewriting for Retrieval-Augmented Large Language Models}.
\newblock \bibinfo{journal}{\emph{CoRR}}  \bibinfo{volume}{abs/2305.14283} (\bibinfo{year}{2023}).
\newblock
\urldef\tempurl%
\url{https://doi.org/10.48550/ARXIV.2305.14283}
\showDOI{\tempurl}
\showeprint[arXiv]{2305.14283}


\bibitem[Malinka et~al\mbox{.}(2023)]%
        {edu-gpt}
\bibfield{author}{\bibinfo{person}{Kamil Malinka}, \bibinfo{person}{Martin Peres{\'{\i}}ni}, \bibinfo{person}{Anton Firc}, \bibinfo{person}{Ondrej Hujnak}, {and} \bibinfo{person}{Filip Janus}.} \bibinfo{year}{2023}\natexlab{}.
\newblock \showarticletitle{On the Educational Impact of ChatGPT: Is Artificial Intelligence Ready to Obtain a University Degree?}. In \bibinfo{booktitle}{\emph{Proceedings of the 2023 Conference on Innovation and Technology in Computer Science Education V. 1, ITiCSE 2023, Turku, Finland, July 7-12, 2023}}, \bibfield{editor}{\bibinfo{person}{Mikko{-}Jussi Laakso}, \bibinfo{person}{Mattia Monga}, \bibinfo{person}{Simon}, {and} \bibinfo{person}{Judithe Sheard}} (Eds.). \bibinfo{publisher}{{ACM}}, \bibinfo{pages}{47--53}.
\newblock
\urldef\tempurl%
\url{https://doi.org/10.1145/3587102.3588827}
\showDOI{\tempurl}


\bibitem[OpenAI(2023)]%
        {gpt4}
\bibfield{author}{\bibinfo{person}{OpenAI}.} \bibinfo{year}{2023}\natexlab{}.
\newblock \showarticletitle{{GPT-4} Technical Report}.
\newblock \bibinfo{journal}{\emph{CoRR}}  \bibinfo{volume}{abs/2303.08774} (\bibinfo{year}{2023}).
\newblock
\urldef\tempurl%
\url{https://doi.org/10.48550/ARXIV.2303.08774}
\showDOI{\tempurl}
\showeprint[arXiv]{2303.08774}


\bibitem[Ouyang et~al\mbox{.}(2022)]%
        {instructgpt}
\bibfield{author}{\bibinfo{person}{Long Ouyang}, \bibinfo{person}{Jeffrey Wu}, \bibinfo{person}{Xu Jiang}, \bibinfo{person}{Diogo Almeida}, \bibinfo{person}{Carroll~L. Wainwright}, \bibinfo{person}{Pamela Mishkin}, \bibinfo{person}{Chong Zhang}, \bibinfo{person}{Sandhini Agarwal}, \bibinfo{person}{Katarina Slama}, \bibinfo{person}{Alex Ray}, \bibinfo{person}{John Schulman}, \bibinfo{person}{Jacob Hilton}, \bibinfo{person}{Fraser Kelton}, \bibinfo{person}{Luke Miller}, \bibinfo{person}{Maddie Simens}, \bibinfo{person}{Amanda Askell}, \bibinfo{person}{Peter Welinder}, \bibinfo{person}{Paul~F. Christiano}, \bibinfo{person}{Jan Leike}, {and} \bibinfo{person}{Ryan Lowe}.} \bibinfo{year}{2022}\natexlab{}.
\newblock \showarticletitle{Training language models to follow instructions with human feedback}. In \bibinfo{booktitle}{\emph{Advances in Neural Information Processing Systems 35: Annual Conference on Neural Information Processing Systems 2022, NeurIPS 2022, New Orleans, LA, USA, November 28 - December 9, 2022}}, \bibfield{editor}{\bibinfo{person}{Sanmi Koyejo}, \bibinfo{person}{S.~Mohamed}, \bibinfo{person}{A.~Agarwal}, \bibinfo{person}{Danielle Belgrave}, \bibinfo{person}{K.~Cho}, {and} \bibinfo{person}{A.~Oh}} (Eds.).
\newblock
\urldef\tempurl%
\url{http://papers.nips.cc/paper\_files/paper/2022/hash/b1efde53be364a73914f58805a001731-Abstract-Conference.html}
\showURL{%
\tempurl}


\bibitem[Papineni et~al\mbox{.}(2002)]%
        {papineni2002bleu}
\bibfield{author}{\bibinfo{person}{Kishore Papineni}, \bibinfo{person}{Salim Roukos}, \bibinfo{person}{Todd Ward}, {and} \bibinfo{person}{Wei{-}Jing Zhu}.} \bibinfo{year}{2002}\natexlab{}.
\newblock \showarticletitle{Bleu: a Method for Automatic Evaluation of Machine Translation}. In \bibinfo{booktitle}{\emph{Proceedings of the 40th Annual Meeting of the Association for Computational Linguistics, July 6-12, 2002, Philadelphia, PA, {USA}}}. \bibinfo{publisher}{{ACL}}, \bibinfo{pages}{311--318}.
\newblock
\urldef\tempurl%
\url{https://doi.org/10.3115/1073083.1073135}
\showDOI{\tempurl}


\bibitem[Shao et~al\mbox{.}(2023)]%
        {shao2023enhancing}
\bibfield{author}{\bibinfo{person}{Zhihong Shao}, \bibinfo{person}{Yeyun Gong}, \bibinfo{person}{Yelong Shen}, \bibinfo{person}{Minlie Huang}, \bibinfo{person}{Nan Duan}, {and} \bibinfo{person}{Weizhu Chen}.} \bibinfo{year}{2023}\natexlab{}.
\newblock \bibinfo{title}{Enhancing Retrieval-Augmented Large Language Models with Iterative Retrieval-Generation Synergy}.
\newblock
\newblock
\showeprint[arxiv]{2305.15294}~[cs.CL]


\bibitem[Shen et~al\mbox{.}(2022)]%
        {shen2022mask}
\bibfield{author}{\bibinfo{person}{Kai Shen}, \bibinfo{person}{Yichong Leng}, \bibinfo{person}{Xu Tan}, \bibinfo{person}{Siliang Tang}, \bibinfo{person}{Yuan Zhang}, \bibinfo{person}{Wenjie Liu}, {and} \bibinfo{person}{Edward Lin}.} \bibinfo{year}{2022}\natexlab{}.
\newblock \showarticletitle{Mask the correct tokens: An embarrassingly simple approach for error correction}.
\newblock \bibinfo{journal}{\emph{arXiv preprint arXiv:2211.13252}} (\bibinfo{year}{2022}).
\newblock


\bibitem[Shi et~al\mbox{.}(2023)]%
        {shi2023replug}
\bibfield{author}{\bibinfo{person}{Weijia Shi}, \bibinfo{person}{Sewon Min}, \bibinfo{person}{Michihiro Yasunaga}, \bibinfo{person}{Minjoon Seo}, \bibinfo{person}{Rich James}, \bibinfo{person}{Mike Lewis}, \bibinfo{person}{Luke Zettlemoyer}, {and} \bibinfo{person}{Wen tau Yih}.} \bibinfo{year}{2023}\natexlab{}.
\newblock \bibinfo{title}{REPLUG: Retrieval-Augmented Black-Box Language Models}.
\newblock
\newblock
\showeprint[arxiv]{2301.12652}~[cs.CL]


\bibitem[Singhal et~al\mbox{.}(2022)]%
        {med-llm}
\bibfield{author}{\bibinfo{person}{Karan Singhal}, \bibinfo{person}{Shekoofeh Azizi}, \bibinfo{person}{Tao Tu}, \bibinfo{person}{S.~Sara Mahdavi}, \bibinfo{person}{Jason Wei}, \bibinfo{person}{Hyung~Won Chung}, \bibinfo{person}{Nathan Scales}, \bibinfo{person}{Ajay~Kumar Tanwani}, \bibinfo{person}{Heather Cole{-}Lewis}, \bibinfo{person}{Stephen Pfohl}, \bibinfo{person}{Perry Payne}, \bibinfo{person}{Martin Seneviratne}, \bibinfo{person}{Paul Gamble}, \bibinfo{person}{Chris Kelly}, \bibinfo{person}{Nathaneal Sch{\"{a}}rli}, \bibinfo{person}{Aakanksha Chowdhery}, \bibinfo{person}{Philip~Andrew Mansfield}, \bibinfo{person}{Blaise~Ag{\"{u}}era y Arcas}, \bibinfo{person}{Dale~R. Webster}, \bibinfo{person}{Gregory~S. Corrado}, \bibinfo{person}{Yossi Matias}, \bibinfo{person}{Katherine Chou}, \bibinfo{person}{Juraj Gottweis}, \bibinfo{person}{Nenad Tomasev}, \bibinfo{person}{Yun Liu}, \bibinfo{person}{Alvin Rajkomar}, \bibinfo{person}{Joelle~K. Barral}, \bibinfo{person}{Christopher Semturs}, \bibinfo{person}{Alan
  Karthikesalingam}, {and} \bibinfo{person}{Vivek Natarajan}.} \bibinfo{year}{2022}\natexlab{}.
\newblock \showarticletitle{Large Language Models Encode Clinical Knowledge}.
\newblock \bibinfo{journal}{\emph{CoRR}}  \bibinfo{volume}{abs/2212.13138} (\bibinfo{year}{2022}).
\newblock
\urldef\tempurl%
\url{https://doi.org/10.48550/ARXIV.2212.13138}
\showDOI{\tempurl}
\showeprint[arXiv]{2212.13138}


\bibitem[Singhal et~al\mbox{.}(2023)]%
        {med-llm2}
\bibfield{author}{\bibinfo{person}{Karan Singhal}, \bibinfo{person}{Tao Tu}, \bibinfo{person}{Juraj Gottweis}, \bibinfo{person}{Rory Sayres}, \bibinfo{person}{Ellery Wulczyn}, \bibinfo{person}{Le Hou}, \bibinfo{person}{Kevin Clark}, \bibinfo{person}{Stephen Pfohl}, \bibinfo{person}{Heather Cole{-}Lewis}, \bibinfo{person}{Darlene Neal}, \bibinfo{person}{Mike Schaekermann}, \bibinfo{person}{Amy Wang}, \bibinfo{person}{Mohamed Amin}, \bibinfo{person}{Sami Lachgar}, \bibinfo{person}{Philip~Andrew Mansfield}, \bibinfo{person}{Sushant Prakash}, \bibinfo{person}{Bradley Green}, \bibinfo{person}{Ewa Dominowska}, \bibinfo{person}{Blaise~Ag{\"{u}}era y Arcas}, \bibinfo{person}{Nenad Tomasev}, \bibinfo{person}{Yun Liu}, \bibinfo{person}{Renee Wong}, \bibinfo{person}{Christopher Semturs}, \bibinfo{person}{S.~Sara Mahdavi}, \bibinfo{person}{Joelle~K. Barral}, \bibinfo{person}{Dale~R. Webster}, \bibinfo{person}{Gregory~S. Corrado}, \bibinfo{person}{Yossi Matias}, \bibinfo{person}{Shekoofeh Azizi}, \bibinfo{person}{Alan
  Karthikesalingam}, {and} \bibinfo{person}{Vivek Natarajan}.} \bibinfo{year}{2023}\natexlab{}.
\newblock \showarticletitle{Towards Expert-Level Medical Question Answering with Large Language Models}.
\newblock \bibinfo{journal}{\emph{CoRR}}  \bibinfo{volume}{abs/2305.09617} (\bibinfo{year}{2023}).
\newblock
\urldef\tempurl%
\url{https://doi.org/10.48550/ARXIV.2305.09617}
\showDOI{\tempurl}
\showeprint[arXiv]{2305.09617}


\bibitem[Sun et~al\mbox{.}(2021)]%
        {ernie}
\bibfield{author}{\bibinfo{person}{Yu Sun}, \bibinfo{person}{Shuohuan Wang}, \bibinfo{person}{Shikun Feng}, \bibinfo{person}{Siyu Ding}, \bibinfo{person}{Chao Pang}, \bibinfo{person}{Junyuan Shang}, \bibinfo{person}{Jiaxiang Liu}, \bibinfo{person}{Xuyi Chen}, \bibinfo{person}{Yanbin Zhao}, \bibinfo{person}{Yuxiang Lu}, \bibinfo{person}{Weixin Liu}, \bibinfo{person}{Zhihua Wu}, \bibinfo{person}{Weibao Gong}, \bibinfo{person}{Jianzhong Liang}, \bibinfo{person}{Zhizhou Shang}, \bibinfo{person}{Peng Sun}, \bibinfo{person}{Wei Liu}, \bibinfo{person}{Xuan Ouyang}, \bibinfo{person}{Dianhai Yu}, \bibinfo{person}{Hao Tian}, \bibinfo{person}{Hua Wu}, {and} \bibinfo{person}{Haifeng Wang}.} \bibinfo{year}{2021}\natexlab{}.
\newblock \showarticletitle{{ERNIE} 3.0: Large-scale Knowledge Enhanced Pre-training for Language Understanding and Generation}.
\newblock \bibinfo{journal}{\emph{CoRR}}  \bibinfo{volume}{abs/2107.02137} (\bibinfo{year}{2021}).
\newblock
\showeprint[arXiv]{2107.02137}
\urldef\tempurl%
\url{https://arxiv.org/abs/2107.02137}
\showURL{%
\tempurl}


\bibitem[Team(2023)]%
        {internlm}
\bibfield{author}{\bibinfo{person}{InternLM Team}.} \bibinfo{year}{2023}\natexlab{}.
\newblock \bibinfo{title}{InternLM: A Multilingual Language Model with Progressively Enhanced Capabilities}.
\newblock \bibinfo{howpublished}{\url{https://github.com/InternLM/InternLM}}.
\newblock


\bibitem[Touvron et~al\mbox{.}(2023)]%
        {llama}
\bibfield{author}{\bibinfo{person}{Hugo Touvron}, \bibinfo{person}{Thibaut Lavril}, \bibinfo{person}{Gautier Izacard}, \bibinfo{person}{Xavier Martinet}, \bibinfo{person}{Marie{-}Anne Lachaux}, \bibinfo{person}{Timoth{\'{e}}e Lacroix}, \bibinfo{person}{Baptiste Rozi{\`{e}}re}, \bibinfo{person}{Naman Goyal}, \bibinfo{person}{Eric Hambro}, \bibinfo{person}{Faisal Azhar}, \bibinfo{person}{Aur{\'{e}}lien Rodriguez}, \bibinfo{person}{Armand Joulin}, \bibinfo{person}{Edouard Grave}, {and} \bibinfo{person}{Guillaume Lample}.} \bibinfo{year}{2023}\natexlab{}.
\newblock \showarticletitle{LLaMA: Open and Efficient Foundation Language Models}.
\newblock \bibinfo{journal}{\emph{CoRR}}  \bibinfo{volume}{abs/2302.13971} (\bibinfo{year}{2023}).
\newblock
\urldef\tempurl%
\url{https://doi.org/10.48550/ARXIV.2302.13971}
\showDOI{\tempurl}
\showeprint[arXiv]{2302.13971}


\bibitem[VoyageAI(2023)]%
        {VoyageAI}
\bibfield{author}{\bibinfo{person}{VoyageAI}.} \bibinfo{year}{2023}\natexlab{}.
\newblock \bibinfo{title}{VoyageAI. Voyage’s embedding models}.
\newblock \bibinfo{howpublished}{\url{https://docs.voyageai.com/embeddings/}}.
\newblock


\bibitem[Wang et~al\mbox{.}(2023)]%
        {wang2023query2doc}
\bibfield{author}{\bibinfo{person}{Liang Wang}, \bibinfo{person}{Nan Yang}, {and} \bibinfo{person}{Furu Wei}.} \bibinfo{year}{2023}\natexlab{}.
\newblock \bibinfo{title}{Query2doc: Query Expansion with Large Language Models}.
\newblock
\newblock
\showeprint[arxiv]{2303.07678}~[cs.IR]


\bibitem[Wu et~al\mbox{.}(2020)]%
        {wu2020biased}
\bibfield{author}{\bibinfo{person}{Yiquan Wu}, \bibinfo{person}{Kun Kuang}, \bibinfo{person}{Yating Zhang}, \bibinfo{person}{Xiaozhong Liu}, \bibinfo{person}{Changlong Sun}, \bibinfo{person}{Jun Xiao}, \bibinfo{person}{Yueting Zhuang}, \bibinfo{person}{Luo Si}, {and} \bibinfo{person}{Fei Wu}.} \bibinfo{year}{2020}\natexlab{}.
\newblock \showarticletitle{De-biased court’s view generation with causality}. In \bibinfo{booktitle}{\emph{Proceedings of the 2020 Conference on Empirical Methods in Natural Language Processing (EMNLP)}}. \bibinfo{pages}{763--780}.
\newblock


\bibitem[Wu et~al\mbox{.}(2023a)]%
        {wu2023focus}
\bibfield{author}{\bibinfo{person}{Yiquan Wu}, \bibinfo{person}{Weiming Lu}, \bibinfo{person}{Yating Zhang}, \bibinfo{person}{Adam Jatowt}, \bibinfo{person}{Jun Feng}, \bibinfo{person}{Changlong Sun}, \bibinfo{person}{Fei Wu}, {and} \bibinfo{person}{Kun Kuang}.} \bibinfo{year}{2023}\natexlab{a}.
\newblock \showarticletitle{Focus-aware response generation in inquiry conversation}. In \bibinfo{booktitle}{\emph{Findings of the Association for Computational Linguistics: ACL 2023}}. \bibinfo{pages}{12585--12599}.
\newblock


\bibitem[Wu et~al\mbox{.}(2023b)]%
        {wu2023precedent}
\bibfield{author}{\bibinfo{person}{Yiquan Wu}, \bibinfo{person}{Siying Zhou}, \bibinfo{person}{Yifei Liu}, \bibinfo{person}{Weiming Lu}, \bibinfo{person}{Xiaozhong Liu}, \bibinfo{person}{Yating Zhang}, \bibinfo{person}{Changlong Sun}, \bibinfo{person}{Fei Wu}, {and} \bibinfo{person}{Kun Kuang}.} \bibinfo{year}{2023}\natexlab{b}.
\newblock \showarticletitle{Precedent-Enhanced Legal Judgment Prediction with LLM and Domain-Model Collaboration}.
\newblock \bibinfo{journal}{\emph{arXiv preprint arXiv:2310.09241}} (\bibinfo{year}{2023}).
\newblock


\bibitem[Xiao et~al\mbox{.}(2023)]%
        {bge_embedding}
\bibfield{author}{\bibinfo{person}{Shitao Xiao}, \bibinfo{person}{Zheng Liu}, \bibinfo{person}{Peitian Zhang}, {and} \bibinfo{person}{Niklas Muennighoff}.} \bibinfo{year}{2023}\natexlab{}.
\newblock \bibinfo{title}{C-Pack: Packaged Resources To Advance General Chinese Embedding}.
\newblock
\newblock
\showeprint[arxiv]{2309.07597}~[cs.CL]


\bibitem[Xu et~al\mbox{.}(2023)]%
        {xu2023recomp}
\bibfield{author}{\bibinfo{person}{Fangyuan Xu}, \bibinfo{person}{Weijia Shi}, {and} \bibinfo{person}{Eunsol Choi}.} \bibinfo{year}{2023}\natexlab{}.
\newblock \bibinfo{title}{RECOMP: Improving Retrieval-Augmented LMs with Compression and Selective Augmentation}.
\newblock
\newblock
\showeprint[arxiv]{2310.04408}~[cs.CL]


\bibitem[Yang et~al\mbox{.}(2023)]%
        {baichuan}
\bibfield{author}{\bibinfo{person}{Aiyuan Yang}, \bibinfo{person}{Bin Xiao}, \bibinfo{person}{Bingning Wang}, \bibinfo{person}{Borong Zhang}, \bibinfo{person}{Ce Bian}, \bibinfo{person}{Chao Yin}, \bibinfo{person}{Chenxu Lv}, \bibinfo{person}{Da Pan}, \bibinfo{person}{Dian Wang}, \bibinfo{person}{Dong Yan}, \bibinfo{person}{Fan Yang}, \bibinfo{person}{Fei Deng}, \bibinfo{person}{Feng Wang}, \bibinfo{person}{Feng Liu}, \bibinfo{person}{Guangwei Ai}, \bibinfo{person}{Guosheng Dong}, \bibinfo{person}{Haizhou Zhao}, \bibinfo{person}{Hang Xu}, \bibinfo{person}{Haoze Sun}, \bibinfo{person}{Hongda Zhang}, \bibinfo{person}{Hui Liu}, \bibinfo{person}{Jiaming Ji}, \bibinfo{person}{Jian Xie}, \bibinfo{person}{Juntao Dai}, \bibinfo{person}{Kun Fang}, \bibinfo{person}{Lei Su}, \bibinfo{person}{Liang Song}, \bibinfo{person}{Lifeng Liu}, \bibinfo{person}{Liyun Ru}, \bibinfo{person}{Luyao Ma}, \bibinfo{person}{Mang Wang}, \bibinfo{person}{Mickel Liu}, \bibinfo{person}{MingAn Lin}, \bibinfo{person}{Nuolan Nie},
  \bibinfo{person}{Peidong Guo}, \bibinfo{person}{Ruiyang Sun}, \bibinfo{person}{Tao Zhang}, \bibinfo{person}{Tianpeng Li}, \bibinfo{person}{Tianyu Li}, \bibinfo{person}{Wei Cheng}, \bibinfo{person}{Weipeng Chen}, \bibinfo{person}{Xiangrong Zeng}, \bibinfo{person}{Xiaochuan Wang}, \bibinfo{person}{Xiaoxi Chen}, \bibinfo{person}{Xin Men}, \bibinfo{person}{Xin Yu}, \bibinfo{person}{Xuehai Pan}, \bibinfo{person}{Yanjun Shen}, \bibinfo{person}{Yiding Wang}, \bibinfo{person}{Yiyu Li}, \bibinfo{person}{Youxin Jiang}, \bibinfo{person}{Yuchen Gao}, \bibinfo{person}{Yupeng Zhang}, \bibinfo{person}{Zenan Zhou}, {and} \bibinfo{person}{Zhiying Wu}.} \bibinfo{year}{2023}\natexlab{}.
\newblock \showarticletitle{Baichuan 2: Open Large-scale Language Models}.
\newblock \bibinfo{journal}{\emph{CoRR}}  \bibinfo{volume}{abs/2309.10305} (\bibinfo{year}{2023}).
\newblock
\urldef\tempurl%
\url{https://doi.org/10.48550/ARXIV.2309.10305}
\showDOI{\tempurl}
\showeprint[arXiv]{2309.10305}


\bibitem[Yue et~al\mbox{.}(2023b)]%
        {fedjudge}
\bibfield{author}{\bibinfo{person}{Linan Yue}, \bibinfo{person}{Qi Liu}, \bibinfo{person}{Yichao Du}, \bibinfo{person}{Weibo Gao}, \bibinfo{person}{Ye Liu}, {and} \bibinfo{person}{Fangzhou Yao}.} \bibinfo{year}{2023}\natexlab{b}.
\newblock \showarticletitle{FedJudge: Federated Legal Large Language Model}.
\newblock \bibinfo{journal}{\emph{CoRR}}  \bibinfo{volume}{abs/2309.08173} (\bibinfo{year}{2023}).
\newblock
\urldef\tempurl%
\url{https://doi.org/10.48550/ARXIV.2309.08173}
\showDOI{\tempurl}
\showeprint[arXiv]{2309.08173}


\bibitem[Yue et~al\mbox{.}(2023a)]%
        {lawllm}
\bibfield{author}{\bibinfo{person}{Shengbin Yue}, \bibinfo{person}{Wei Chen}, \bibinfo{person}{Siyuan Wang}, \bibinfo{person}{Bingxuan Li}, \bibinfo{person}{Chenchen Shen}, \bibinfo{person}{Shujun Liu}, \bibinfo{person}{Yuxuan Zhou}, \bibinfo{person}{Yao Xiao}, \bibinfo{person}{Song Yun}, \bibinfo{person}{Xuanjing Huang}, {and} \bibinfo{person}{Zhongyu Wei}.} \bibinfo{year}{2023}\natexlab{a}.
\newblock \showarticletitle{DISC-LawLLM: Fine-tuning Large Language Models for Intelligent Legal Services}.
\newblock \bibinfo{journal}{\emph{CoRR}}  \bibinfo{volume}{abs/2309.11325} (\bibinfo{year}{2023}).
\newblock
\urldef\tempurl%
\url{https://doi.org/10.48550/ARXIV.2309.11325}
\showDOI{\tempurl}
\showeprint[arXiv]{2309.11325}


\bibitem[Zeng et~al\mbox{.}(2023)]%
        {glm}
\bibfield{author}{\bibinfo{person}{Aohan Zeng}, \bibinfo{person}{Xiao Liu}, \bibinfo{person}{Zhengxiao Du}, \bibinfo{person}{Zihan Wang}, \bibinfo{person}{Hanyu Lai}, \bibinfo{person}{Ming Ding}, \bibinfo{person}{Zhuoyi Yang}, \bibinfo{person}{Yifan Xu}, \bibinfo{person}{Wendi Zheng}, \bibinfo{person}{Xiao Xia}, \bibinfo{person}{Weng~Lam Tam}, \bibinfo{person}{Zixuan Ma}, \bibinfo{person}{Yufei Xue}, \bibinfo{person}{Jidong Zhai}, \bibinfo{person}{Wenguang Chen}, \bibinfo{person}{Zhiyuan Liu}, \bibinfo{person}{Peng Zhang}, \bibinfo{person}{Yuxiao Dong}, {and} \bibinfo{person}{Jie Tang}.} \bibinfo{year}{2023}\natexlab{}.
\newblock \showarticletitle{{GLM-130B:} An Open Bilingual Pre-trained Model}. In \bibinfo{booktitle}{\emph{The Eleventh International Conference on Learning Representations, {ICLR} 2023, Kigali, Rwanda, May 1-5, 2023}}. \bibinfo{publisher}{OpenReview.net}.
\newblock
\urldef\tempurl%
\url{https://openreview.net/pdf?id=-Aw0rrrPUF}
\showURL{%
\tempurl}


\bibitem[Zhang et~al\mbox{.}(2024)]%
        {zhang2024plad}
\bibfield{author}{\bibinfo{person}{Rongzhi Zhang}, \bibinfo{person}{Jiaming Shen}, \bibinfo{person}{Tianqi Liu}, \bibinfo{person}{Haorui Wang}, \bibinfo{person}{Zhen Qin}, \bibinfo{person}{Feng Han}, \bibinfo{person}{Jialu Liu}, \bibinfo{person}{Simon Baumgartner}, \bibinfo{person}{Michael Bendersky}, {and} \bibinfo{person}{Chao Zhang}.} \bibinfo{year}{2024}\natexlab{}.
\newblock \bibinfo{title}{PLaD: Preference-based Large Language Model Distillation with Pseudo-Preference Pairs}.
\newblock
\newblock
\showeprint[arxiv]{2406.02886}~[cs.CL]


\bibitem[Zhao et~al\mbox{.}(2023)]%
        {domain-survey}
\bibfield{author}{\bibinfo{person}{Xujiang Zhao}, \bibinfo{person}{Jiaying Lu}, \bibinfo{person}{Chengyuan Deng}, \bibinfo{person}{Can Zheng}, \bibinfo{person}{Junxiang Wang}, \bibinfo{person}{Tanmoy Chowdhury}, \bibinfo{person}{Li Yun}, \bibinfo{person}{Hejie Cui}, \bibinfo{person}{Zhang Xuchao}, \bibinfo{person}{Tianjiao Zhao}, {et~al\mbox{.}}} \bibinfo{year}{2023}\natexlab{}.
\newblock \showarticletitle{Domain specialization as the key to make large language models disruptive: A comprehensive survey}.
\newblock \bibinfo{journal}{\emph{arXiv preprint arXiv:2305.18703}} (\bibinfo{year}{2023}).
\newblock


\bibitem[Zhou et~al\mbox{.}(2023)]%
        {lima}
\bibfield{author}{\bibinfo{person}{Chunting Zhou}, \bibinfo{person}{Pengfei Liu}, \bibinfo{person}{Puxin Xu}, \bibinfo{person}{Srini Iyer}, \bibinfo{person}{Jiao Sun}, \bibinfo{person}{Yuning Mao}, \bibinfo{person}{Xuezhe Ma}, \bibinfo{person}{Avia Efrat}, \bibinfo{person}{Ping Yu}, \bibinfo{person}{Lili Yu}, \bibinfo{person}{Susan Zhang}, \bibinfo{person}{Gargi Ghosh}, \bibinfo{person}{Mike Lewis}, \bibinfo{person}{Luke Zettlemoyer}, {and} \bibinfo{person}{Omer Levy}.} \bibinfo{year}{2023}\natexlab{}.
\newblock \showarticletitle{{LIMA:} Less Is More for Alignment}.
\newblock \bibinfo{journal}{\emph{CoRR}}  \bibinfo{volume}{abs/2305.11206} (\bibinfo{year}{2023}).
\newblock
\urldef\tempurl%
\url{https://doi.org/10.48550/ARXIV.2305.11206}
\showDOI{\tempurl}
\showeprint[arXiv]{2305.11206}


\bibitem[Zhou et~al\mbox{.}(2022)]%
        {zhou2022similar}
\bibfield{author}{\bibinfo{person}{Siying Zhou}, \bibinfo{person}{Yifei Liu}, \bibinfo{person}{Yiquan Wu}, \bibinfo{person}{Kun Kuang}, \bibinfo{person}{Chunyan Zheng}, {and} \bibinfo{person}{Fei Wu}.} \bibinfo{year}{2022}\natexlab{}.
\newblock \showarticletitle{Similar case based prison term prediction}. In \bibinfo{booktitle}{\emph{CAAI International Conference on Artificial Intelligence}}. Springer, \bibinfo{pages}{284--297}.
\newblock


\end{thebibliography}
